\documentclass[a4paper,fleqn]{cas-sc}

\usepackage[authoryear,longnamesfirst]{natbib}
\usepackage{subcaption}
\usepackage{lineno}
\usepackage{hyperref}
\usepackage{amsmath,amssymb}
\usepackage{graphicx}
\usepackage{booktabs}
\usepackage{microtype}
\usepackage{microtype}
\usepackage{graphicx}
\usepackage{subcaption}
\usepackage{booktabs} % for professional tables
\usepackage{multirow}
\usepackage{tabularx}
\usepackage{float}
\usepackage{placeins}
\usepackage{caption}
\usepackage{hyperref}
\usepackage{svg}
\def\tsc#1{\csdef{#1}{\textsc{\lowercase{#1}}\xspace}}
\tsc{WGM}
\tsc{QE}
\tsc{EP}
\tsc{PMS}
\tsc{BEC}
\tsc{DE}
\begin{document}
\let\WriteBookmarks\relax
\def\floatpagepagefraction{1}
\def\textpagefraction{.001}
\shorttitle{Agile Reinforcement Learning through Separable Neural Architecture}
\shortauthors{Rajib Mostakim, Reza T. Batley and Sourav Saha}
%\begin{frontmatter}

\title[mode=title]{Agile Reinforcement Learning through Separable Neural Architecture and Applications}

\author[a]{Rajib Mostakim}

\author[b]{Reza T. Batley}

\author[b]{Sourav Saha\corref{cor1}}
\ead{souravsaha@vt.edu}

\cortext[cor1]{Corresponding author}

\address[a]{Department of Mechanical Engineering, Bangladesh University of Engineering and Technology, Dhaka, Bangladesh}

\address[b]{Kevin T. Crofton Department of Aerospace and Ocean Engineering, Virginia Polytechnic Institute and State University, Blacksburg, VA, USA}

\begin{abstract}
Deep reinforcement learning (RL) is increasingly deployed in resource-constrained environments, yet the go-to function approximators - multilayer perceptrons (MLPs) - are often parameter-inefficient due to an imperfect inductive bias for the smooth structure of many value functions. This mismatch can also hinder sample efficiency and slow policy learning in this capacity-limited regime. Although model compression techniques exist, they operate post-hoc and do not improve learning efficiency. Recent spline-based separable architectures such as Kolmogorov-Arnold Networks (KANs) have been shown to offer parameter efficiency but are widely reported to exhibit significant computational overhead, especially at scale. In seeking to address these limitations, this work introduces SPAN (\textbf{SP}line-based \textbf{A}daptive \textbf{N}etworks), a novel function approximation approach to RL. SPAN adapts the KHRONOS framework by integrating a learnable preprocessing layer. SPAN is evaluated across discrete (PPO) and high-dimensional continuous (SAC) control tasks, offline settings (Minari/D4RL) and a real-world datacenter HVAC control application. Empirical results demonstrate that SPAN achieves a \textbf{30-50\% improvement in sample efficiency} and \textbf{1.3-9 times higher success rates} across benchmarks compared to MLP baselines. Despite incurring a \textbf{per-step evaluation overhead of 1.2-1.8 times} , SPAN's superior convergence reliability yields an expected total \textbf{training cost 1.3-6.3 times lower} than MLP baselines when accounting for convergence failures. In the HVAC application, SPAN \textbf{reduces energy consumption in 9 of 12 months} relative to MLP while simultaneously achieving a \textbf{ 1.1-3.4 times reduction in thermal comfort violations} across the evaluation year, demonstrating generalization to real-world engineering control. Furthermore, SPAN demonstrates superior anytime performance and robustness to hyperparameter variations, suggesting it as a viable, high-performance alternative for learning intrinsically efficient policies in resource-limited settings.
\end{abstract}

%\begin{graphicalabstract}
%\includegraphics{figs/cas-grabs.pdf}
%\end{graphicalabstract}

%\begin{highlights}
%\item Research highlights item 1
%\item Research highlights item 2
%\item Research highlights item 3
%\end{highlights}

\begin{keywords}
B-spline neural networks \sep separable architecture \sep actor-critic reinforcement learning \sep sample efficiency \sep parameter-efficient learning \sep continuous control

\end{keywords}
%\end{frontmatter}

\maketitle

\section{Introduction}

Deep reinforcement learning has achieved remarkable success across diverse domains including game playing  \citep{mnih2015humanlevel, silver2017mastering}, robotic manipulation \citep{levine2016endtoend, kalashnikov2018qt}, autonomous navigation \citep{kahn2017plato, tai2017virtual}, and industrial optimization \citep{mao2016resource}. These achievements fundamentally rely on neural function approximation to represent policies and value functions. The dominant approach employs multilayer perceptrons (MLPs), chosen for their universal approximation properties \citep{hornik1989multilayer}.

However, deploying RL in real-world applications faces critical resource constraints. Physical systems incur real costs per interaction and have limited samples \citep{dulacarnold2019challengesrealworld}, edge devices impose strict memory budgets \citep{li2020edgeai}, and production environments frequently face training interruptions \citep{zilberstein1996operational}. Under these constraints, MLPs remain limited: their globally connected structure distributes representational capacity uniformly across the input space, regardless of where the value function actually varies. This mismatch between architecture and problem structure leads to high sample complexity \citep{kakade2003sample} and poor convergence reliability precisely when resources are most limited.

A natural alternative is architectures whose inductive bias is explicitly aligned with the structure of the functions being approximated. RL value functions and policies exhibit local smoothness: formally, under Lipschitz-continuous MDP dynamics, the value function inherits Lipschitz continuity \citep{rachelson2010}, and the policy gradient landscape is smooth with respect to policy parameters \citep{pirotta2015}. KHRONOS with B-spline basis functions at its core is architecturally matched to this structure: each basis function activates only in a bounded region of the input space, concentrating representational capacity where it is needed. Spline-based architectures such as Kolmogorov-Arnold Networks \citep{liu2025kan} offer local support but suffer from dense edge-wise evaluation that results in prohibitive computational overhead \citep{kich2024kan}, limiting practical adoption in online RL.

This work introduces \textbf{SPAN} (\textbf{SP}line-based \textbf{A}daptive \textbf{N}etworks), a parameter-efficient function approximation architecture for resource-constrained RL. SPAN adapts KHRONOS framework \citep{reza}, previously validated for supervised regression in scientific computing, to the online RL setting through a learnable preprocessing layer that projects arbitrary, non-stationary RL observations into the bounded domain required by the KHRONOS. By leveraging a rank-$M$ tensor decomposition, SPAN reduces parameter complexity from exponential to linear in input dimension, overcoming the curse of dimensionality that limits grid-based spline approaches. SPAN is designed as a drop-in replacement for MLPs in standard actor-critic frameworks, leaving training objectives and update rules unchanged.

The contributions of this work are threefold. \textbf{First}, we present an architectural adaptation of the KHRONOS framework to reinforcement learning, introducing a learnable preprocessing layer that bridges the gap between the stationary normalised inputs assumed by the original framework and the non-stationary unbounded observations of RL environments. \textbf{Second}, we provide a theoretical grounding connecting SPAN's local support property to the Lipschitz structure of RL value functions and policies, establishing an RL-specific justification for the separable spline inductive bias. \textbf{Third}, we empirically validate SPAN across online (PPO, SAC) offline (IQL) and an application in RL settings under strict parameter budgets, demonstrating a 30-50\% improvement in sample efficiency, 1.3-9 times higher success rates, and an expected total training cost 1.3-6.3 times lower than MLP baselines when accounting for convergence reliability.

The remainder of this article is organized as follows. Section \ref{sec:related} reviews prior work in RL and the recent emergence of spline-based architectures in scientific computing. Section \ref{sec:background} establishes RL preliminaries . Section \ref{sec:method} details the SPAN architecture. Section \ref{sec:experiments} presents empirical evaluations across online and offline control benchmarks. Section \ref{sec:application} demonstrates SPAN's advantage in a HVAC application, and Section \ref{sec:discussion} analyzes the structural inductive bias and discusses current limitations and future direction followed by concluding remarks in Section \ref{sec:conclusion}.

\section{Related Work}
\label{sec:related}

Modern deep RL predominantly employs multilayer perceptrons (MLPs) for policy and value function approximation \citep{mnih2015humanlevel, lillicrap2019continuous, schulman2017ppo, haarnoja2018sac}. While justified by universal approximation guarantees \citep{hornik1989multilayer}, standard MLPs typically require hundreds of hidden units to represent smooth control landscapes accurately. To improve optimization stability, prior research has explored activation functions such as Swish and GELU \citep{ramachandran2017swish, hendrycks2023gelu}, as well as architectural stabilizers including layer normalization and residual connections \citep{ba2016layer, he2015deepresidual}. Spectral normalization has also been applied directly to actor-critic networks to bound the Lipschitz constant of the policy and value function \citep{bjorck2022}. While these techniques improve optimization stability, they impose smoothness as a constraint on an otherwise globally-connected architecture: every weight remains coupled to every input dimension, and the constraint bounds the \emph{rate} of change of the learned function without concentrating representational capacity locally. Consequently, these approaches preserve the uniformly distributed capacity of the MLP, which often leads to sample inefficiency in learning.

A complementary line of work imposes locality directly through the choice of basis function. Radial basis function (RBF) networks \citep{sutton2018reinforcement} and Fourier basis methods \citep{konidaris2011} were widely used in early value function approximation, and Gaussian process approaches provided a Bayesian treatment of smooth value functions \citep{engel2003bayes, engel2005reinforcement}. RBF networks provide genuine local support but require a number of basis centres that grows exponentially with input dimension \citep{carse1996}, while Fourier features have global support and, as \citep{konidaris2011} note, struggle precisely where the value function requires local representation. Gaussian process methods scale cubically in the number of observations \citep{rasmussen2006gaussian}, making them intractable for online RL.

A substantial body of work focuses on reducing the size of neural networks via pruning \citep{han2016deepcompression}, quantization \citep{jacob2017quantization}, or knowledge distillation \citep{hinton2015distillingknowledge}. However, these techniques generally operate as \emph{post hoc} optimization steps on over-parameterized models, primarily targeting inference latency rather than training efficiency. Similarly, factorized networks \citep{sainath2013low} and tensor decompositions \citep{novikov2015tensordecompose} reduce parameter counts by imposing low rank structures on weight matrices, but retain global connectivity. In contrast, SPAN targets \textit{intrinsic learning efficiency}, reducing the number of environment interactions required to learn a performant policy directly under fixed, low capacity constraints, without expensive pre-training or compression phases.

Recently, Kolmogorov-Arnold Networks (KANs) \citep{liu2025kan} have emerged as a spline-based alternative to MLPs for reducing parameter counts while preserving expressive power. In offline reinforcement learning, KANs can match MLP performance with fewer parameters, but pure KAN architectures often underperform compared to hybrid KAN--MLP models and incur higher training costs \citep{guo2024kanoffline}. Similarly, KANs have been applied to online continuous control tasks \citep{kich2024kan}, reporting comparable performance to MLPs with reduced memory usage. However, this study highlighted a significant computational bottleneck: the dense evaluation of B-splines on every network edge resulted in training times approximately nine times slower than standard MLPs, and performance stability varied significantly across tasks.

Parallel to KANs, the KHRONOS framework \citep{reza} introduced a separable tensor product architecture for scientific computing. The framework has been validated across diverse scientific domains, including multi-fidelity aerodynamic prediction \citep{apurbaaero}, inverse material design \citep{rezajanus}, predictive and generative intelligence \citep{batley2026snaprimitive}, world model application \citep{batley2026sna}, smooth token embeddings \citep{batley2026leviathan}, fatigue life estimation and application in additive manufacturing \citep{rajib, Park2025}. However, while these works demonstrate the efficacy of the architecture for supervised regression, generative intelligence, and world modeling, its application to the dynamic, sequential decision making challenges of Reinforcement Learning remains unexplored.

This work introduces SPAN to bridge this gap. Rather than using a raw KHRONOS architecture, SPAN incorporates a learnable preprocessing layer that projects unstructured, non-stationary observations into the bounded domain required by KHRONOS, enabling separable tensor-product representations to operate within online RL training loops. Unlike the smoothness-as-constraint approaches discussed above, SPAN's local support provides genuine spatial locality of representation: each basis function activates only in a bounded region of state space, concentrating capacity where the value function and policy actually vary. Section~\ref{sec:background} formalizes the connection between this property and the Lipschitz structure of RL value functions established by \citep{rachelson2010} and \citep{pirotta2015}. Furthermore, this study provides a comprehensive evaluation isolating sample efficiency and anytime performance metrics critical for deployment in resource-constrained settings, rather than solely focusing on final asymptotic performance.

\section{Theoretical Background} \label{sec:background}
\subsection{Reinforcement Learning Preliminaries}
The standard reinforcement learning framework is formalized as a Markov Decision Process (MDP) defined by the tuple $(\mathcal{S}, \mathcal{A}, P, r, \gamma)$, \citep{sutton2018reinforcement} where $\mathcal{S}$ is the state space, $\mathcal{A}$ is the action space, $P: \mathcal{S} \times \mathcal{A} \times \mathcal{S} \rightarrow [0,1]$ is the transition probability function, $r: \mathcal{S} \times \mathcal{A} \rightarrow \mathbb{R}$ is the reward function, and $\gamma \in [0,1)$ is the discount factor.

The agent's behavior is determined by a policy $\pi: \mathcal{S} \rightarrow \mathcal{P}(\mathcal{A})$, which maps states to probability distributions over actions. The objective is to find a policy that maximizes the expected cumulative discounted reward \citep{sutton2000policy}:

$$J(\pi) = \mathbb{E}_{\tau \sim \pi}\left[\sum_{t=0}^{\infty} \gamma^t r(s_t, a_t)\right]$$

where $\tau = (s_0, a_0, s_1, a_1, \ldots)$ denotes a trajectory sampled by executing policy $\pi$.

Value-based and policy-based approaches differ in whether they learn an explicit value function ($V^\pi(s)$), an action-value ($Q^\pi(s,a)$), or a direct parametric policy ($\pi_\theta(a|s)$). In continuous or large discrete state/action spaces, exact representation of value functions or policies is intractable. Modern deep reinforcement learning employs parameterized function approximators $Q_\theta(s,a)$ and $\pi_\phi(a|s)$ to represent these functions. 

\subsection{Reinforcement Learning Algorithms}
\textbf{Soft Actor-Critic (SAC)} is an off-policy actor-critic algorithm for continuous control that optimizes a maximum entropy objective \citep{haarnoja2018sac,haarnoja2019sacapps}. By explicitly encouraging stochasticity in the policy, SAC promotes effective exploration. The algorithm employs a stochastic policy, twin Q-functions to reduce overestimation bias, and slowly updated target networks for stable learning. Policy and value functions are learned jointly from off policy data, and the entropy temperature can be automatically adjusted to balance exploration and reward maximization.

\textbf{Proximal Policy Optimization (PPO) }is an on-policy policy gradient method that improves training stability by restricting the magnitude of policy updates \citep{schulman2017ppo}. PPO uses advantage estimates to guide policy improvement while preventing destructive updates through a clipped objective. In practice, advantages are commonly estimated using Generalized Advantage Estimation (GAE) \citep{schulman2018gae}. PPO further incorporates a value function loss and an entropy bonus.

\textbf{Implicit Q-Learning (IQL)} is an offline reinforcement learning algorithm designed to avoid explicit behavior policy estimation \citep{kostrikov2021iql}. IQL learns a Q-function via asymmetric value regression, encouraging conservatism by penalizing overestimation of unseen actions. Policy learning is performed by advantage weighted regression, where actions with higher estimated advantages are assigned higher likelihood under the learned policy. This implicit constraint enables stable learning from fixed datasets without requiring importance sampling or explicit policy constraints, making IQL effective in offline regimes.

\subsection{Structural Properties of RL Value Functions} \label{sec:smoothness}

The architectural design of SPAN is motivated by a structural property of value functions and policies in physically grounded control tasks: \emph{local smoothness}. We formalize this property below and establish its connection to function approximators with local support.

\textbf{Lipschitz continuity of the value function.} Consider an MDP whose transition dynamics $P$ and reward function $r$ are Lipschitz continuous with respect to the state, i.e., there exist constants $L_P, L_r$ such that for all $a \in \mathcal{A}$ and $s, s' \in \mathcal{S}$,
$$
\mathcal{W}(P(\cdot|s,a), P(\cdot|s',a)) \leq L_P \lVert s - s' \rVert, \qquad |r(s,a) - r(s',a)| \leq L_r \lVert s - s' \rVert,
$$
where $\mathcal{W}$ denotes the Wasserstein distance between transition distributions. \citep{rachelson2010} show that under these conditions, the optimal value function $V^*(s)$ is itself Lipschitz continuous:
$$
|V^*(s) - V^*(s')| \leq L_V \lVert s - s' \rVert,
$$
for some constant $L_V$ depending on $L_P$, $L_r$, and $\gamma$. In words, states that are close under the MDP's dynamics have correspondingly close values - the value function cannot change arbitrarily fast across nearby states.

\textbf{Influence radius.} A direct consequence of this result, also established by \citep{rachelson2010}, is the existence of an \emph{influence radius}: for each state $s$, there exists a neighborhood $\mathcal{B}(s,\rho) \subset \mathcal{S}$ within which the optimal action remains constant. This locality property emerges directly from the sequential decision structure of the MDP and the Bellman recursion, a property of the underlying decision problem itself.

\textbf{Smoothness of the policy optimization landscape.} \citep{pirotta2015} extend this result to the policy optimization setting. Adopting $L_\pi$-Lipschitz continuity of the policy as a foundational assumption nearby states produce nearby action distributions, $\mathcal{W}(\pi(\cdot|s), \pi(\cdot|s')) \leq L_\pi \lVert s-s'\rVert$  they show that under Lipschitz MDP dynamics, the expected return $J(\pi)$ and its gradient $\nabla_\theta J(\pi)$ are themselves Lipschitz continuous with respect to the policy parameters $\theta$. The optimization landscape therefore contains no sudden discontinuities or cliffs: small parameter perturbations produce correspondingly small changes in both performance and its gradient.

\textbf{Architectural alignment with KHRONOS local support.} These results motivate a function approximator whose representational capacity is structured around spatial locality. KHRONOS's basis function $B_i^{(k)}(x_p)$ is non-zero only for $x_p$ within a bounded neighborhood of its associated knot formally, $B_i^{(k)}(x_p) = 0$ for $x_p \notin \Omega_i \subset [0,1]$, where $\Omega_i$ has width proportional to the local knot spacing. Consequently, for an input $x \in [0,1]^d$, a gradient update derived from a transition at the corresponding state contributes only to basis functions $B_i^{(k)}$ for which $x \in \Omega_i$ that is, only to weights associated with the local neighborhood of that state in the input domain. This is a structural property of the representation itself: capacity is concentrated where the agent's experience is concentrated, and updates from one region of the input space do not perturb the representation in distant regions. This is in contrast to globally-connected architectures, including those constrained via spectral normalization (Section~\ref{sec:related}), where every parameter remains coupled to every input dimension and a gradient update from any state continues to influence the function's value everywhere. The local support of KHRONOS instead realizes the influence radius property of \citep{rachelson2010} \emph{by construction}: the spatial extent of a basis function's support plays the role of the influence radius, and the separable tensor-product structure (Section~\ref{sec:method}) extends this locality to the full $d$-dimensional input space without incurring the exponential scaling of grid-based local methods.

\textbf{Scope of the smoothness assumption.} The Lipschitz conditions underlying these results require that nearby states yield nearby transition distributions and rewards. This holds naturally in the dense-reward, ODE governed control tasks where dynamics evolve continuously and reward signals vary smoothly with state. The conditions are substantially weakened in sparse-reward, long-horizon settings, where the value function may be near-constant over large regions of state space before changing sharply near a goal.

\section{Method: SPAN for Reinforcement Learning} \label{sec:method}

\subsection{Architectural Foundation}

SPAN (SPline-based Adaptive Networks) builds upon KHRONOS architectures for function approximation. For completeness, the KHRONOS (Kernel-Expansion Hierarchy for Reduced-Order, Neural-Optimized Surrogates) formulation \citep{reza} is briefly reviewed before introducing SPAN's RL-specific adaptations.

Given input $\mathbf{x} \in [0,1]^d$, KHRONOS constructs $M$ separable modes via per-dimension B-spline basis expansions followed by tensor products:
$$\mathcal{M}_j(\mathbf{x}) = \prod_{p=1}^{d} \sum_{i=1}^{N_p} w_{p,j,i} B_i^{(k)}(x_p)$$
where $B_i^{(k)}$ are B-spline basis functions of degree $k$, $N_p$ is the number of basis functions per-dimension, and $w_{p,j,i}$ are learnable weights. The final output combines all modes through a linear head:
$$f(\mathbf{x}) = \mathbf{W} \begin{bmatrix} \mathcal{M}_1(\mathbf{x}) \\ \vdots \\ \mathcal{M}_M(\mathbf{x}) \end{bmatrix} + \mathbf{b}$$
This separable structure reduces parameter complexity from $\mathcal{O}(N^d)$ (full tensor grid) to $\mathcal{O}(MdN)$ (linear in dimension).

\subsection{SPAN}

While KHRONOS operates on normalized inputs $[0,1]^d$, RL environments yield observations with arbitrary ranges and distributions. A preprocessing layer is therefore required that guarantees outputs in $[0,1]^d$ by construction without relying on batch statistics or pre-specified input bounds, both of which are unavailable under the non-stationary observation distributions of online RL. Table~\ref{tab:preprocessing} compares candidate normalization schemes against this constraint.

\FloatBarrier
\begin{table}[H]
\centering
\small
\caption{Comparison of normalization schemes for mapping inputs to the KHRONOS domain $[0,1]^d$. Only Dense + Sigmoid guarantees compliance by construction.}
\label{tab:preprocessing}
\begin{tabular}{llp{10cm}}
\toprule
\textbf{Option} & \textbf{Output Range} & \textbf{Application to SPAN} \\
\midrule
BatchNorm &
$(-\infty,+\infty)$ &
Does not enforce $[0,1]$; introduces batch-statistics dependency that may destabilize early RL training. \\
LayerNorm &
$(-\infty,+\infty)$ &
Same range issue; also destroys relative scale across input dimensions. \\
Min-Max Norm &
$[0,1]$ &
Requires known input bounds; RL input ranges are unknown and shift non-stationarily during training. \\
Dense + Sigmoid (Ours) &
$(0,1)\subset[0,1]$ &
Enforces domain by construction; learns normalization adaptively without batch statistics or known input bounds. \\
\bottomrule
\end{tabular}
\end{table}

SPAN therefore adopts a single fully-connected layer with sigmoid activation as its preprocessing layer:
$$\mathbf{z} = \sigma(\mathbf{W}_{\text{pre}} \mathbf{s} + \mathbf{b}_{\text{pre}})$$
where $\mathbf{s} \in \mathbb{R}^d$ is the environment observation,
$\mathbf{W}_{\text{pre}} \in \mathbb{R}^{d \times d}$,
$\mathbf{b}_{\text{pre}} \in \mathbb{R}^d$ are the weight matrix and bias vector, and $\sigma(\cdot)$ is the elementwise sigmoid. This adds only $d^2 + d$ parameters negligible relativeto the KHRONOS core while providing three benefits beyond domain enforcement: it standardizes input scales across dimensions, and enables feature recombination before nonlinear spline processing.

\textbf{A Koopman operator perspective.} Beyond the functional justification, the Dense + Sigmoid layer has a mechanistic motivation. The preprocessing layer learns to map raw, potentially entangled environment observations $s \in \mathbb{R}^{d}$ into a latent space $z \in (0,1)^{d}$ in which the dynamics admit a simpler, separable representation. This bears a direct conceptual resemblance to Koopman operator theory \citep{lusch2018koopman, brunton2016sindy} in which nonlinear dynamical systems are lifted into an observable space where the evolution operator becomes approximately linear and separable. The Dense layer learns to disentangle and recombine raw input streams into a low-rank latent space; the Sigmoid projects this into the domain required by the spline basis. Together, they constitute a learned Koopman-style lifting map tailored to the RL setting.

\begin{figure}
    \centering
    \includegraphics[width=0.8\linewidth]{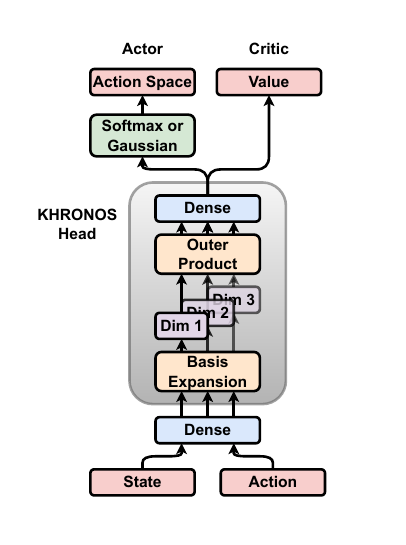}
    \caption{SPAN architecture for actor-critic RL. Both actor and critic use
    identical architectures with separate parameters. The output head produces
    softmax logits for discrete actions or Gaussian parameters for continuous
    actions.}
    \label{fig:span_architecture}
\end{figure}

The full SPAN forward pass composes preprocessing and tensor-product splines
(Figure~\ref{fig:span_architecture}):
$$\text{SPAN}(\mathbf{s};\,\boldsymbol{\Theta}) = \mathbf{W}
\begin{bmatrix} \mathcal{M}_1(\mathbf{z}) \\ \vdots \\ \mathcal{M}_M(\mathbf{z})
\end{bmatrix} + \mathbf{b}, \qquad \mathbf{z} = \sigma(\mathbf{W}_{\text{pre}}
\mathbf{s} + \mathbf{b}_{\text{pre}})$$
where $\boldsymbol{\Theta} = \{\mathbf{W}_{\text{pre}}, \mathbf{b}_{\text{pre}},
\{w_{p,j,i}\}, \mathbf{W}, \mathbf{b}\}$ contains all learnable parameters.

\subsection{Integration with Reinforcement Learning Algorithms}

SPAN serves as a drop-in replacement for MLPs in standard actor-critic frameworks. We detail integration with SAC (continuous control) and PPO (discrete control) and IQL(offline RL).

\subsubsection{Soft Actor-Critic (SAC)}

\textbf{Actor (Gaussian Policy):} For continuous actions, SPAN outputs mean and log-standard-deviation:
$$[\mu_\theta(s); \log \sigma_\theta(s)] = \text{SPAN}_\theta^{\text{actor}}(s)$$
where SPAN$_\theta^{\text{actor}}: \mathbb{R}^d \rightarrow \mathbb{R}^{2A}$ with $A$ the action dimension. Actions are sampled via reparameterization $a = \mu_\theta(s) + \sigma_\theta(s) \odot \epsilon$ with $\epsilon \sim \mathcal{N}(0, I)$, followed by $\tanh$ squashing for bounded action spaces.

\textbf{Twin Critics:} Each Q-function maps concatenated state-action pairs to scalar values:
$$Q_{\theta_i}(s,a) = \text{SPAN}_{\theta_i}^{\text{critic}}([s; a]), \quad i \in \{1,2\}$$
where SPAN$_{\theta_i}^{\text{critic}}: \mathbb{R}^{d+A} \rightarrow \mathbb{R}$. The SAC training loop proceeds unchanged with SPAN as the function approximator.

\subsubsection{Proximal Policy Optimization (PPO)}

\textbf{Policy:} For discrete actions, SPAN outputs logits over $A$ actions:
$$\pi_\theta(a|s) = \text{Categorical}(\text{softmax}(\text{SPAN}_\theta^{\text{actor}}(s)))$$
where SPAN$_\theta^{\text{actor}}: \mathbb{R}^d \rightarrow \mathbb{R}^A$.

\textbf{Value Function:} The critic estimates state values via SPAN$_\phi^{\text{critic}}: \mathbb{R}^d \rightarrow \mathbb{R}$:
$$V_\phi(s) = \text{SPAN}_\phi^{\text{critic}}(s)$$
GAE computation and PPO's clipped objective remain identical to standard implementations.

\subsubsection{Implicit Q-Learning (IQL)} IQL requires three distinct function approximators: a value function $V(s)$, a Q-function $Q(s,a)$, and a policy $\pi(a|s)$. SPAN replaces the MLP in all three roles:
\begin{gather*}
    V_\psi(s) = \text{SPAN}_\psi^{\text{value}}(s), \quad 
    Q_\theta(s,a) = \text{SPAN}_\theta^{\text{critic}}([s; a]) \\
    \pi_\phi(a|s) = \mathcal{N}(\mu_\phi(s), \Sigma_\phi(s)) \quad \text{via} \quad \text{SPAN}_\phi^{\text{actor}}(s)
\end{gather*} where inputs to the Q-network are concatenated state-action pairs $[s; a]$.

\section{Experimental Setup and Results} \label{sec:experiments}

SPAN is evaluated against standard MLP baselines across diverse RL domains
using PPO for Classic Control (CartPole, Acrobot) and Box2D (LunarLander),
SAC for continuous control (Hopper, InvertedPendulum, HalfCheetah), and IQL
for offline manipulation (Adroit Hand). To assess generalization beyond synthetic benchmarks, SPAN is additionally evaluated on a real-world building energy optimization task using the Sinergym datacenter environment \citep{campoy2025sinergym}, detailed in Section \ref{sec:application}. The performance of online RL is assessed through two complementary benchmarks: \textit{sample efficiency} and \textit{anytime performance} that directly address key practical constraints in real-world RL deployment. For offline RL, as online interaction is prohibited during training, the final performance result is reported. 

\textbf{Sample Efficiency} is quantified as the number of environment interaction steps required for an agent to reach specific performance milestones, measured at 25\%, 50\%, 75\%, 95\%, and 100\% of an expert-level score.  This metric is particularly important in robotics and real-world control, where data collection is costly and often performed directly on physical systems \citep{dulacarnold2019challengesrealworld}. Reducing the number of environment interactions lowers deployment costs, mechanical wear, and wall-clock time. Moreover, resource-constrained platforms such as mobile robots and embedded devices impose strict memory limits, making improvements through model scaling impractical. Given the high variance and sensitivity of deep reinforcement learning algorithms \citep{henderson2019deeprlmatters}, methods that achieve target performance with fewer samples can substantially accelerate experimentation and deployment.

\textbf{Anytime Performance}  is the agent performance at multiple checkpoints during training: 10\% 25\%, 50\%, 75\%, 95\%and 100\% of the total training budget steps. At each checkpoint, mean return with standard deviation is measured. This metric is crucial for production environments where training may be preempted due to strict time constraints, budget exhaustion, or resource contention in shared clusters. Anytime performance quantifies the quality of policies available at arbitrary stopping points, enabling practitioners to optimize the trade-off between performance and computational cost.

Together, sample efficiency and anytime performance provide a holistic measure of learning efficiency, capturing both the speed to reach target performance and the evolution of policy quality throughout training. Under real-world constraints such as limited interactions, or fixed model capacities, both metrics are critical. Sample efficiency results, including success rates and median steps to each threshold, are presented in this section alongside raw evaluation curves showing learning trajectories. Anytime performance results are provided in Appendix~\ref{app:anytime}. Training details, algorithm specific hyperparameters, and total training parameter counts are reported in Appendix~\ref{app:training_details}. Complete ablation studies are reported in Appendix~\ref{app:ablation}, statistical evaluation details and computational cost analysis are reported on \ref{app:statistical} and hardware Specifications are reported in Appendix~\ref{app:hardware}.

\subsection{Online Reinforcement Learning Results}

\begin{figure}[!htbp]
\centering
\begin{subfigure}[b]{0.32\textwidth}
    \includegraphics[width=\textwidth]{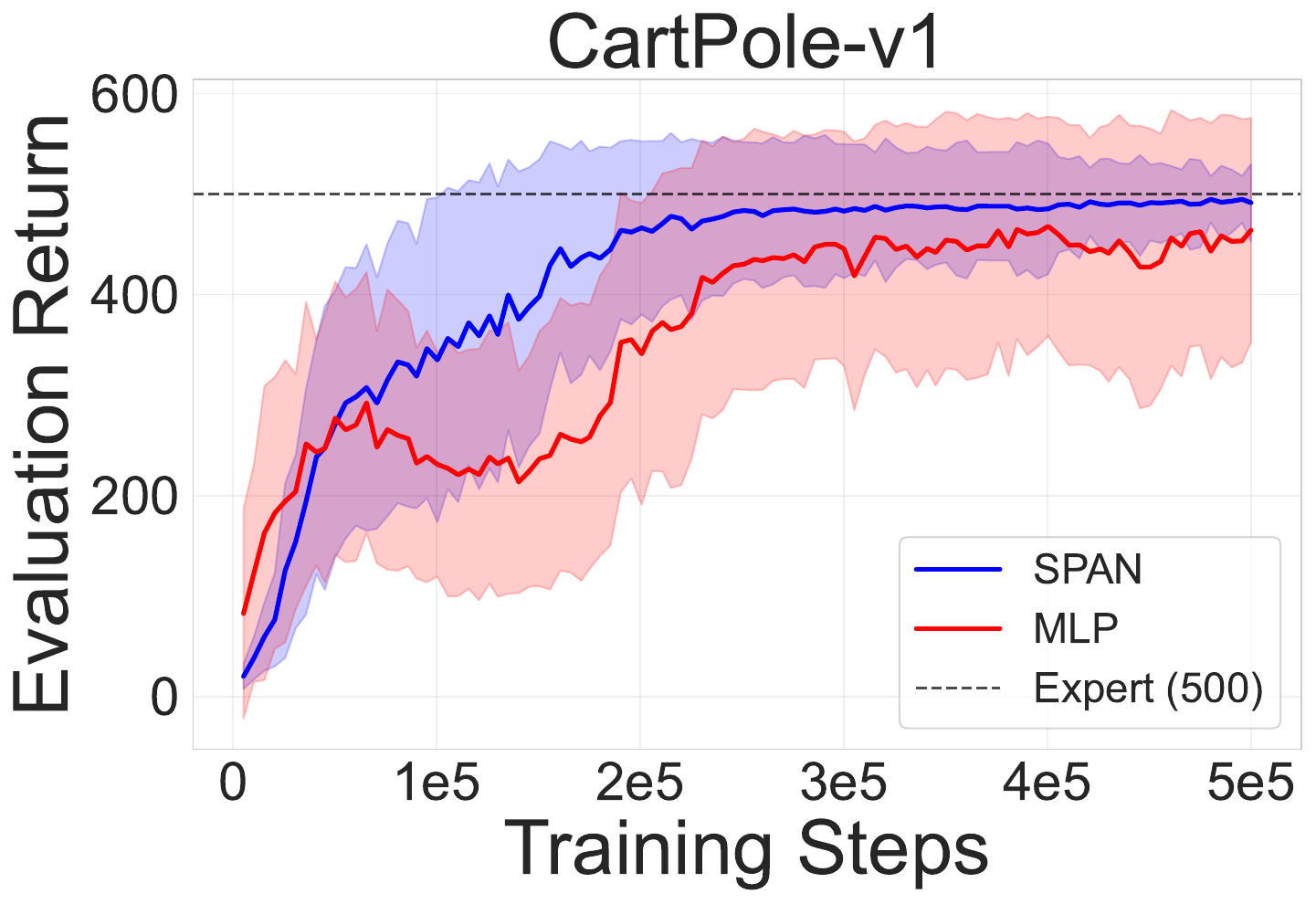}
    %\caption{CartPole}
    \label{fig:cartpole_raw}
\end{subfigure}
\hfill
\begin{subfigure}[b]{0.32\textwidth}
    \includegraphics[width=\textwidth]{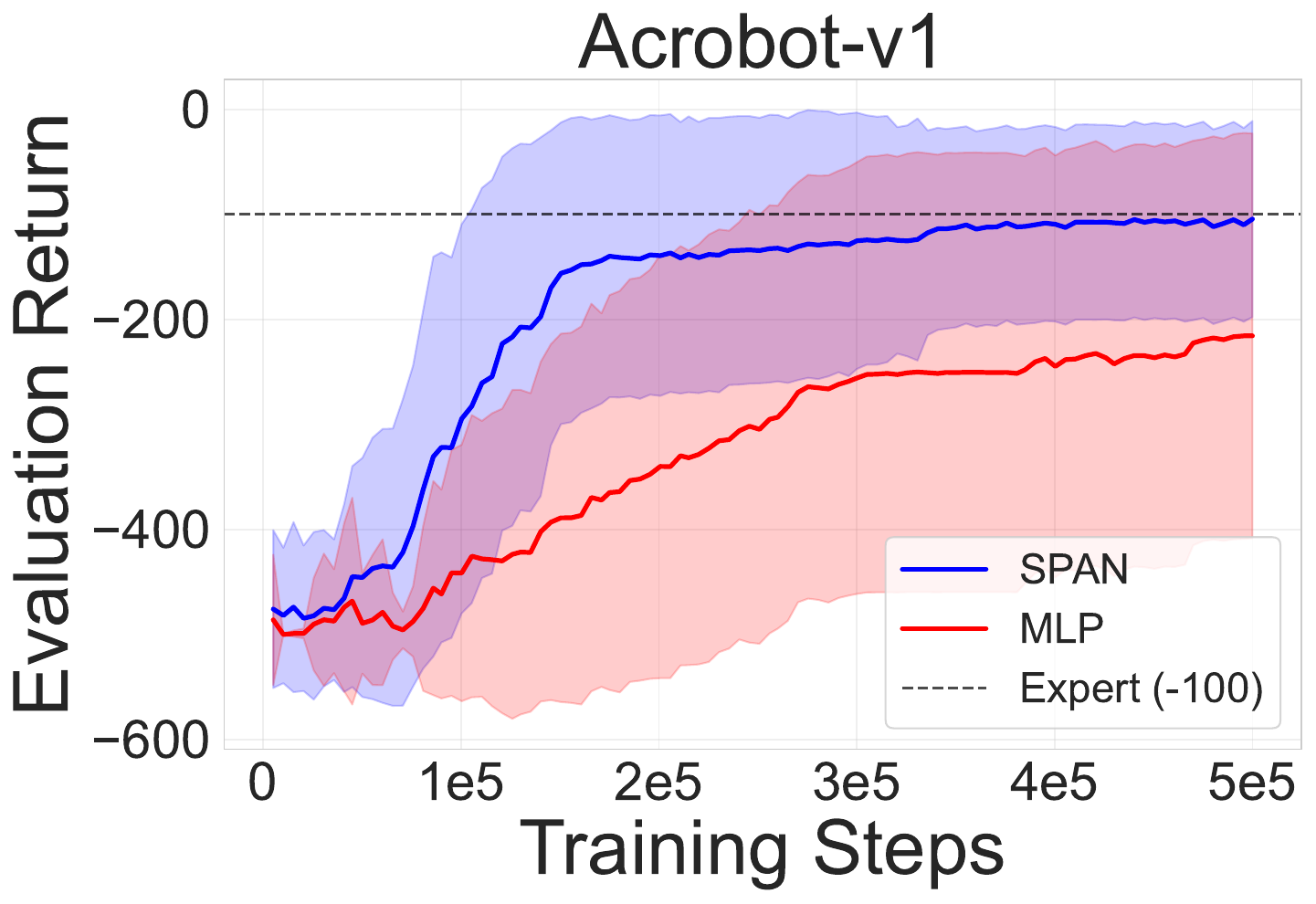}
    %\caption{Acrobot}
    \label{fig:acrobot_raw}
\end{subfigure}
\hfill
\begin{subfigure}[b]{0.32\textwidth}
    \includegraphics[width=\textwidth]{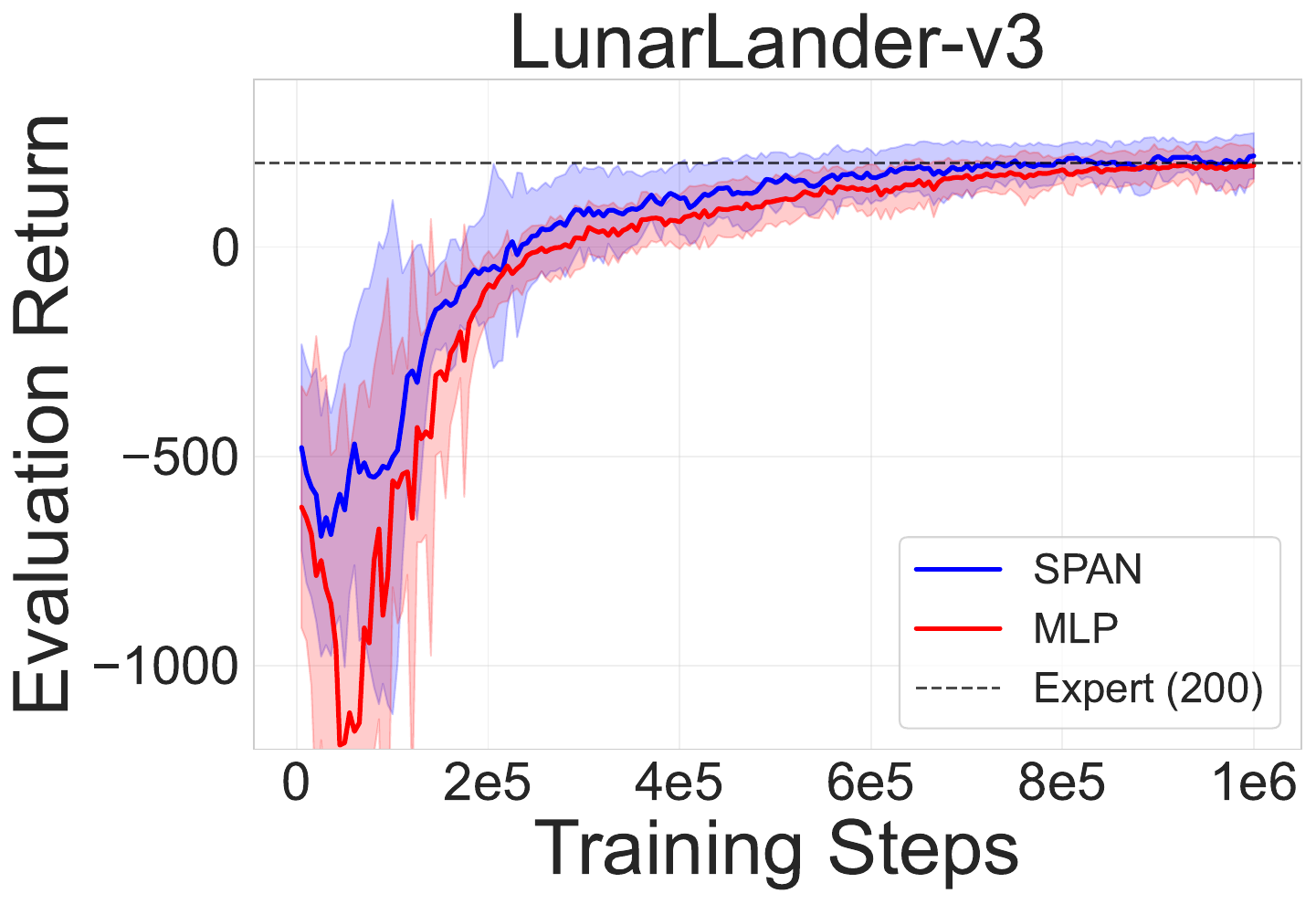}
    %\caption{LunarLander}
    \label{fig:lunarlander_raw}
\end{subfigure}
\caption{Reward curves for Classic Control and Box2D environments showing mean performance ± 1 standard deviation across 20 seeds.}
\label{fig:classic_curves}
\end{figure}

\textbf{Classic Control}: Two standard Gymnasium discrete control environments CartPole and Acrobot and one Box2D environment, LunarLander, are evaluated using PPO \citep{schulman2017ppo}. CartPole involves stabilizing a pole on a moving cart, Acrobot requires swinging a two-link underactuated pendulum to a target height and LunarLander tasks the agent with navigating a spacecraft to a designated landing pad using main and side thrusters.
Figure~\ref{fig:classic_curves} reports learning curves showing mean episodic return with $\pm$ one standard deviation across 20 random seeds. Sample efficiency results are summarized in Table~\ref{tab:classic_efficiency}, while anytime performance is reported in Appendix A.

Across all Classic Control environments, SPAN consistently demonstrates faster learning and reduced variance compared to the MLP baseline. In CartPole, SPAN reaches 50\% of expert performance in a median of 73k environment steps, compared to 86k for MLP, and achieves a 100\% success rate versus 95\% for MLP. At the 100\% performance threshold, SPAN solves the task in 201k steps with a 95\% success rate, whereas MLP requires 300k steps and achieves only 75\% success.

LunarLander exhibits moderate but consistent improvements. SPAN reaches 50\% expert performance in 283k steps, compared to 476k for MLP, corresponding to a 40\% reduction in required interactions, with both methods achieving perfect success rates. At the 100\% threshold, SPAN maintains a 90\% success rate while solving in 618k steps, whereas MLP drops to a 60\% success rate and requires 810k steps. Acrobot, characterized by sparse rewards and negative returns, further highlights SPAN’s robustness. At the 50\% threshold (corresponding to a score of -200), SPAN achieves a 95\% success rate and solves the task in 105k steps, compared to MLP’s 65\% success rate and 175k steps. This approximately 40\% speedup persists at the 100\% threshold, where SPAN maintains a 90\% success rate while MLP again drops to 60\%.

\begin{table*}[t]
\centering
\small
\caption{Sample efficiency on Classic Control environments. Median steps to threshold (lower is better), with success rates in parentheses. SPAN achieves faster learning and higher success rates across all tasks.}
\label{tab:classic_efficiency}
\begin{tabular*}{\textwidth}{@{\extracolsep{\fill}}llrrrrr}
\toprule
Environment & Method & 25\% & 50\% & 75\% & 95\% & 100\% \\
\midrule
\multirow{2}{*}{CartPole} & SPAN & 36k (100\%) & \textbf{73k (100\%)} & \textbf{100k (95\%)} & \textbf{145k (95\%)} & \textbf{201k (95\%)} \\
& MLP & \textbf{20k (95\%)} & 86k (95\%) & 201k (95\%) & 218k (90\%) & 300k (75\%) \\
\midrule
\multirow{2}{*}{LunarLander} & SPAN & \textbf{273k (100\%)} & \textbf{283k (100\%)} & \textbf{430k (95\%)} & \textbf{560k (95\%)} & \textbf{618k (90\%)} \\
& MLP & 348k (100\%) & 476k (100\%) & 621k (95\%) & 751k (75\%) & 810k (60\%) \\
\midrule
\multirow{2}{*}{Acrobot} & SPAN & \textbf{95k (100\%)} & \textbf{105k (100\%)} & \textbf{111k (95\%)} & \textbf{130k (90\%)} & \textbf{145k (90\%)} \\
& MLP & 175k (100\%) & 175k (70\%) & 198k (60\%) & 220k (60\%) & 228k (60\%) \\
\bottomrule
\end{tabular*}
\end{table*}

\textbf{MuJoCo Continuous Control}: SPAN is evaluated on four MuJoCo continuous control environments: InvertedPendulum, Hopper, Walker2d and HalfCheetah using Soft Actor-Critic (SAC) \citep{haarnoja2018sac}. InvertedPendulum requires balancing a single link pendulum in an upright position by applying continuous horizontal forces to a sliding cart. Hopper is a locomotion task in which a planar one legged robot must learn to hop forward efficiently while maintaining balance. Walker2d features a planar bipedal robot that must coordinate its two legs to walk forward as fast as possible without losing its balance and falling over. HalfCheetah models a planar bipedal agent with articulated joints that must learn to run forward at high speed.

\begin{figure}[!hbtp]
\centering
\begin{subfigure}[b]{0.48\textwidth}
    \centering
    \includegraphics[width=\textwidth]{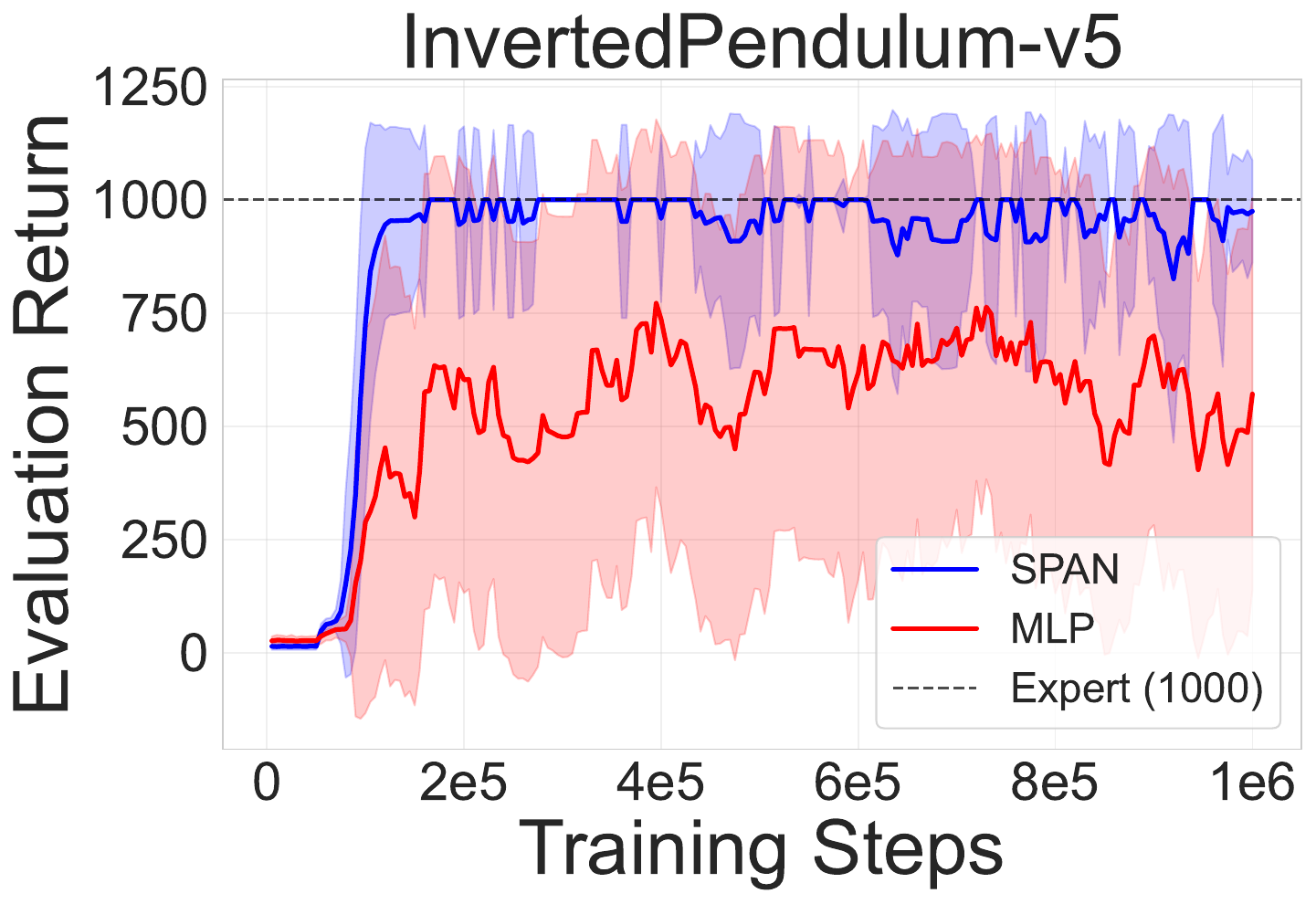}
    %\caption{InvertedPendulum}
    \label{fig:invertedpendulum_raw}
\end{subfigure}
\hfill
\begin{subfigure}[b]{0.48\textwidth}
    \centering
    \includegraphics[width=\textwidth]{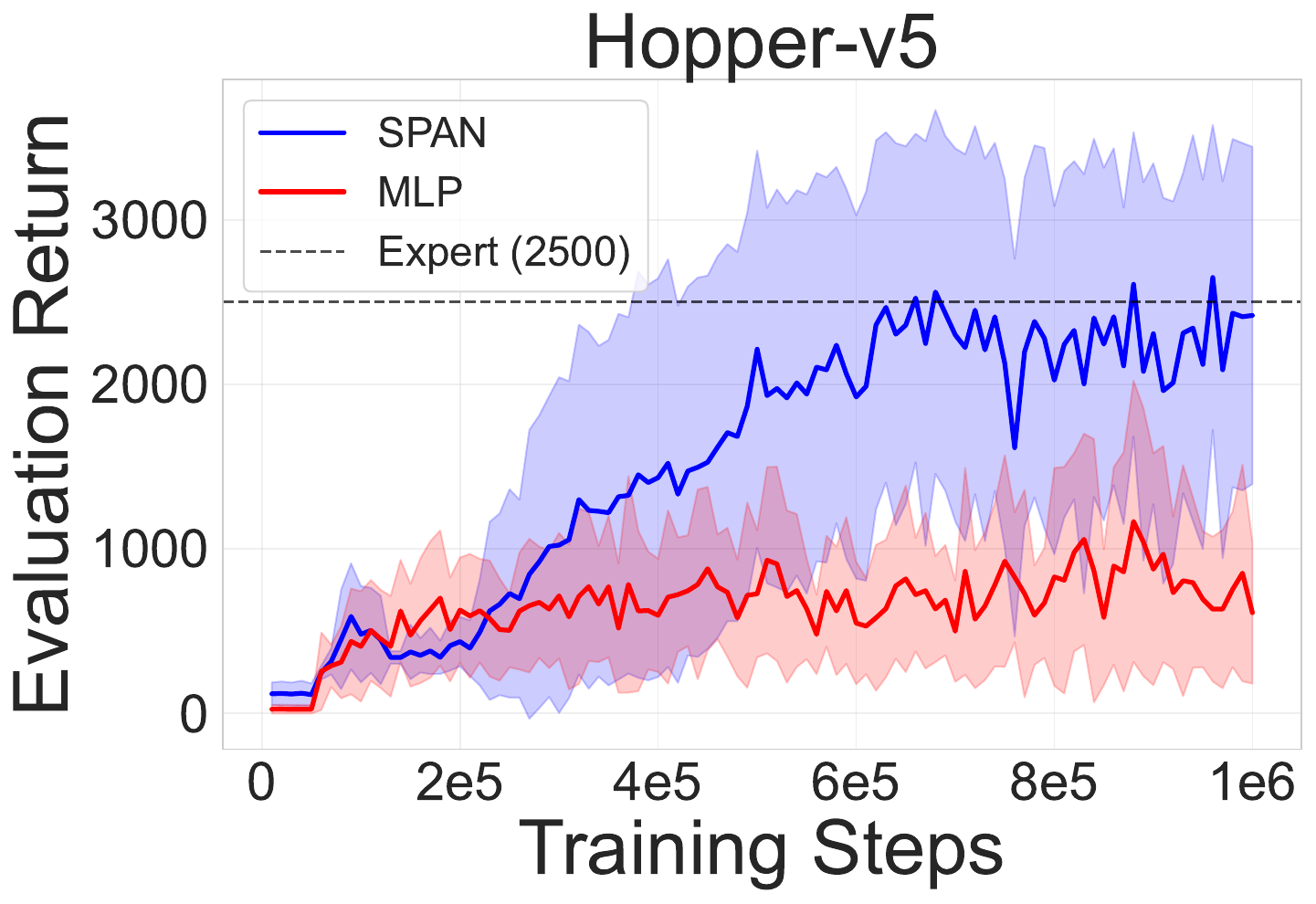}
    %\caption{Hopper}
    \label{fig:hopper_raw}
\end{subfigure}
\caption{Learning curves for InvertedPendulum and Hopper continuous control environments showing mean episodic return $\pm$ one standard deviation across 20 random seeds.}
\label{fig:raw_curves_IP_HP}
\end{figure}

\begin{figure}[!hbtp]
\centering
\begin{subfigure}[b]{0.48\textwidth}
    \centering
    \includegraphics[width=\textwidth]{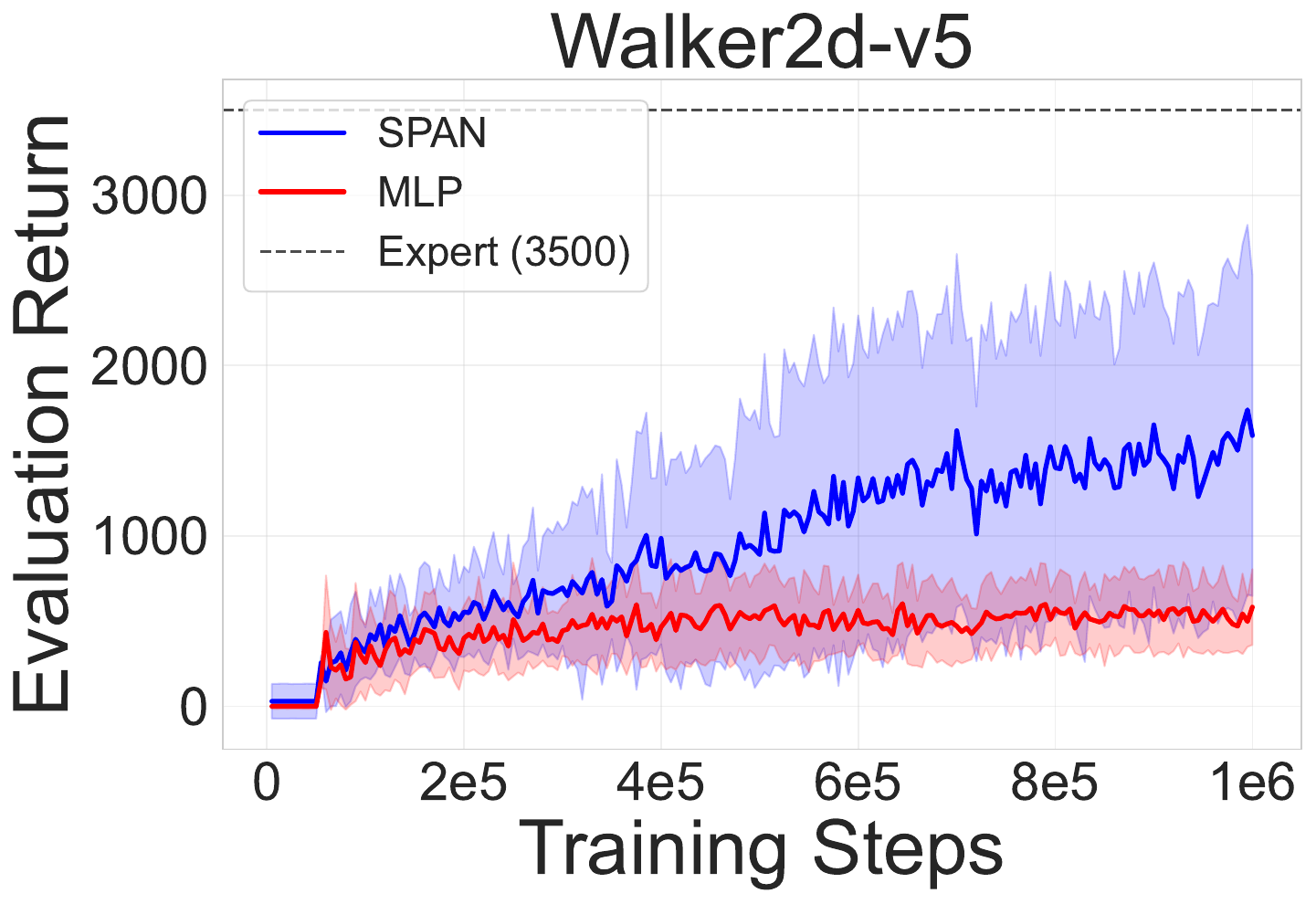}
    %\caption{}
    %\label{fig:walker2d_raw}
\end{subfigure}
\hfill
\begin{subfigure}[b]{0.48\textwidth}
    \includegraphics[width=\textwidth]{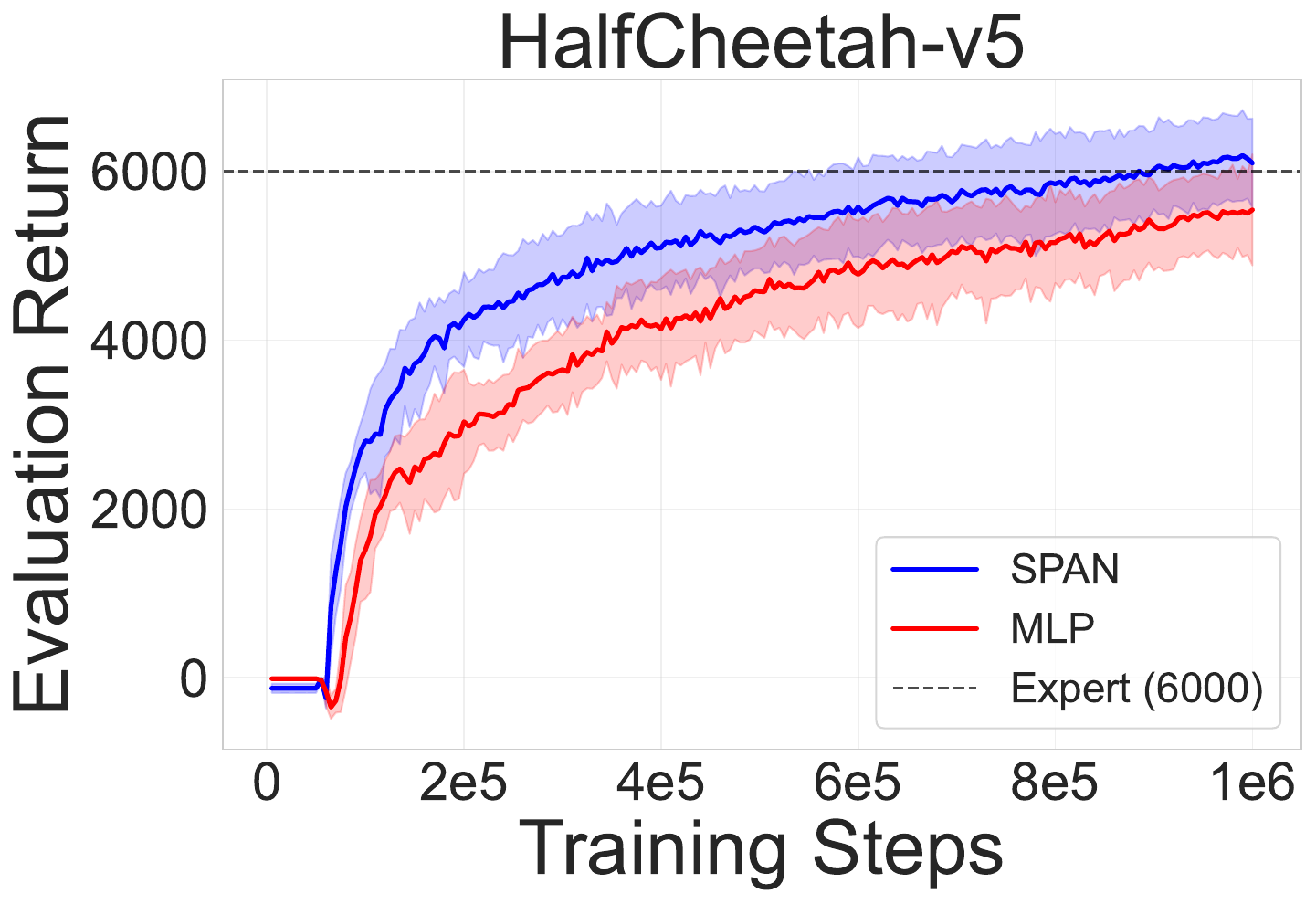}
    %\caption{HalfCheetah}
    %\label{fig:halfcheetah_raw}
\end{subfigure}

\caption{Learning curves for Walker2d and HalfCheetah continuous control environments showing mean episodic return $\pm$ one standard deviation across 20 random seeds.}
\label{fig:raw_curves_W_H}
\end{figure}

\begin{table*}[!hbtp]
\centering
\small
\caption{Sample efficiency on MuJoCo environments. Median steps to threshold (lower is better), with success rates in parentheses. SPAN achieves faster learning and higher success rates across all tasks.}
\label{tab:mujoco_efficiency}
\begin{tabular*}{\textwidth}{@{\extracolsep{\fill}}llrrrrr}
\toprule
Environment & Method & 25\% & 50\% & 75\% & 95\% & 100\% \\
\midrule
\multirow{2}{*}{Hopper} & SPAN & \textbf{310k (95\%)} & \textbf{475k (90\%)} & \textbf{510k (90\%)} & \textbf{495k (80\%)} & \textbf{495k (80\%)} \\
& MLP & 435k (90\%) & 715k (10\%) & - (0\%) & - (0\%) & - (0\%) \\
\midrule
\multirow{2}{*}{InvertedPendulum} & SPAN & \textbf{95k (100\%)} & \textbf{100k (100\%)} & \textbf{100k (100\%)} & \textbf{100k (100\%)} & \textbf{100k (100\%)} \\
& MLP & 178k (100\%) & 178k (100\%) & 178k (100\%) & 178k (100\%) & 178k (100\%) \\
\midrule
\multirow{2}{*}{HalfCheetah} & SPAN & \textbf{75k (100\%)} & \textbf{120k (100\%)} & \textbf{245k (100\%)} & \textbf{590k (90\%)} & \textbf{698k (60\%)} \\
& MLP & 100k (100\%) & 193k (100\%) & 485k (100\%) & 850k (35\%) & 900k (10\%) \\
\midrule
\multirow{2}{*}{Walker2d} & SPAN & \textbf{535k (65\%)} & \textbf{625k (55\%)} & \textbf{653k (30\%)} & - (0\%) & - (0\%) \\
& MLP & 590k (20\%) & - (0\%) & - (0\%) & - (0\%) & - (0\%) \\
\bottomrule
\end{tabular*}
\end{table*}

Figure~\ref{fig:raw_curves_IP_HP} and \ref{fig:raw_curves_W_H} reports learning curves showing mean episodic return with $\pm$ one standard deviation across 20 random seeds. Table~\ref{tab:mujoco_efficiency} reports sample efficiency results across multiple performance thresholds.

Hopper highlights SPAN’s advantage over MLP. At the 50\% performance threshold, SPAN achieves a 90\% success rate with a median solve time of 475k environment steps, while MLP reaches only 10\% success at 715k steps. At higher thresholds (75\% and above), MLP fails to achieve successful solutions across all seeds, whereas SPAN maintains success rates between 80\% and 90\%.

InvertedPendulum exhibits that SPAN reaches the 75\% performance threshold in 100k steps, compared to 178k steps for MLP, corresponding to a 44\% reduction in required interactions.  SPAN consistently converges approximately 2.2$\times$ faster across all thresholds. On HalfCheetah, both methods achieve strong final performance, but SPAN consistently improves sample efficiency. At the 50\% threshold, SPAN solves the task in 120k steps compared to 193k steps for MLP (38\% faster). This advantage increases at higher thresholds: 245k versus 485k steps at 75\% (49\% faster), and at the 95\% threshold SPAN achieves a 90\% success rate compared to 35\% for MLP, with median solve times of 590k and 850k steps, respectively.

Walker2d highlights SPAN's advantage in tasks requiring sustained locomotion stability. At the 25\% performance threshold, SPAN achieves a 65\% success rate with a median solve time of 535k steps, while MLP reaches only 20\% success at 590k steps. At higher thresholds (50\% and above), MLP fails entirely across all seeds, whereas SPAN continues to find solutions at the 50\% threshold (55\% success, 625k steps) and 75\% threshold (30\% success, 653k steps), demonstrating a meaningful capability gap under strict parameter budgets. Both architectures fail to achieve 95\% and 100\% thresholds at this parameter budget. 

A notable discrepancy exists between the raw evaluation curves and the sample efficiency tables that warrants clarification. For InvertedPendulum, the raw learning curves suggest an MLP success rate of approximately 50\%, yet the sample efficiency table reports 100\% success across all thresholds. For Hopper, the MLP curve barely rises above zero for most seeds, yet the 25\% threshold entry reports a 90\% success rate. These apparent contradictions arise from how sample efficiency is defined in this work. To guard against spurious threshold crossings, we adopt a \textit{sustained solving} criterion: a seed is considered to have reached a threshold only if the target return is achieved across thirty evaluation episodes for five consecutive evaluation intervals. The reported \textit{steps-to-threshold} is the first step at which this sustained criterion is met(median taken for twenty seeds), and the \textit{success rate} reflects the fraction of seeds that satisfy it at least once during training. The 100\% success rate for MLP on InvertedPendulum therefore indicates that every seed achieves sustained threshold performance at some point during training, yet the raw curves reveal that many seeds subsequently suffer catastrophic forgetting and fail to maintain that level. Similarly, the 90\% success rate for MLP on Hopper at the 25\% threshold reflects seeds that briefly sustain a low-level hopping gait early in training, even though the median curve never progresses beyond this rudimentary behavior. This distinction highlights an important limitation of threshold-based metrics when used in isolation and further motivates the complementary use of IQM and performance profiles, which capture the \textit{distribution} of final performance rather than a single crossing event.

% ============================================================
% Section 5.2 — Statistical Robustness Analysis
% ============================================================

\subsection{Statistical Robustness Analysis}
\label{sec:statistical}

To provide principled performance comparisons beyond raw learning curves, we adopt the evaluation methodology of \citep{agarwal2022}, computing the Interquartile Mean (IQM) with 95\% bootstrap confidence intervals across 20 seeds, performance profiles over normalised score thresholds, and Welch's $t$-tests with Cohen's $d$ effect sizes.

\begin{figure}[h]
    \centering
    \begin{subfigure}[t]{0.32\textwidth}
        \includegraphics[width=\linewidth]{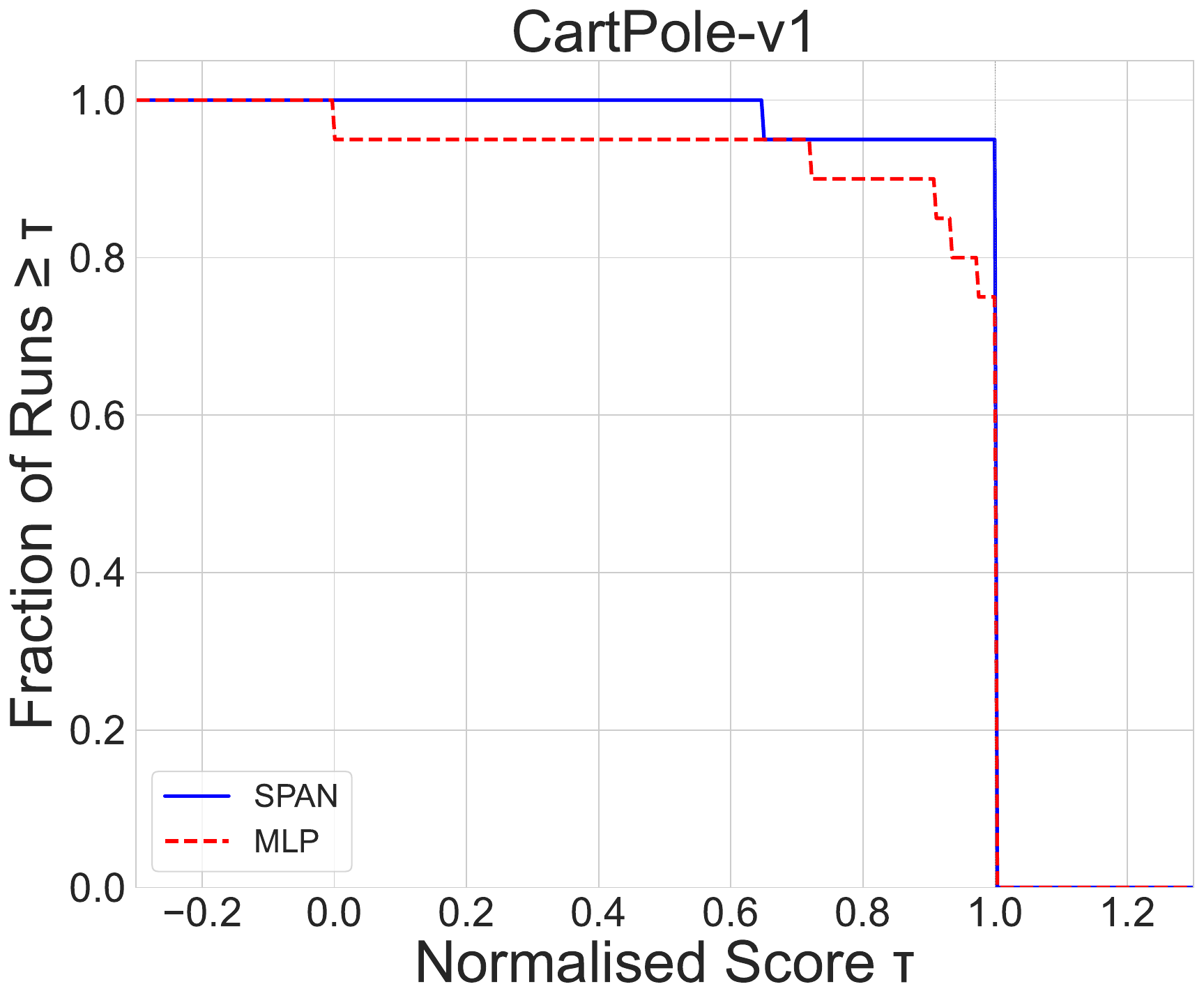}
    \end{subfigure}
    \hfill
    \begin{subfigure}[t]{0.32\textwidth}
        \includegraphics[width=\linewidth]{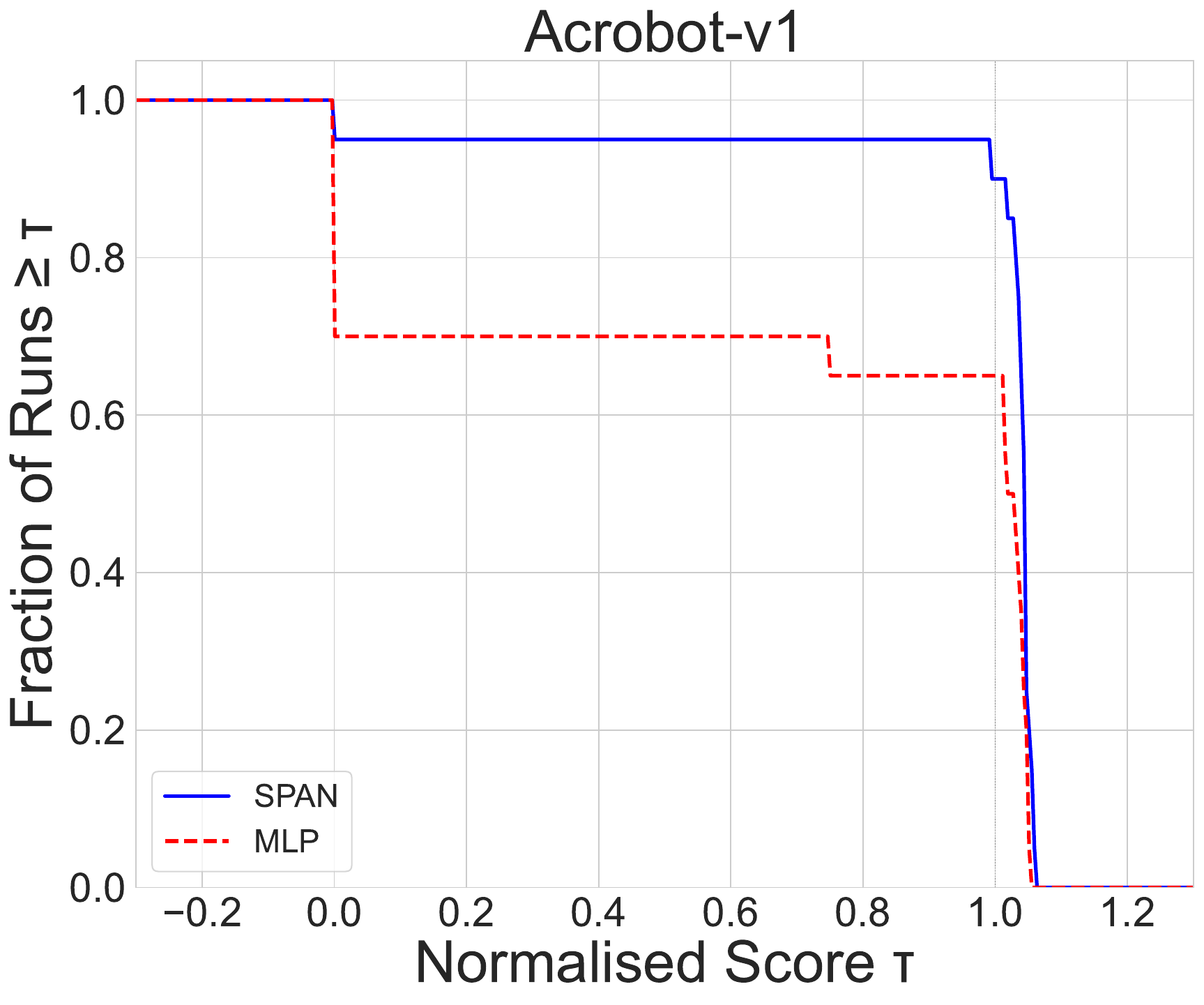}
    \end{subfigure}
    \hfill
    \begin{subfigure}[t]{0.32\textwidth}
        \includegraphics[width=\linewidth]{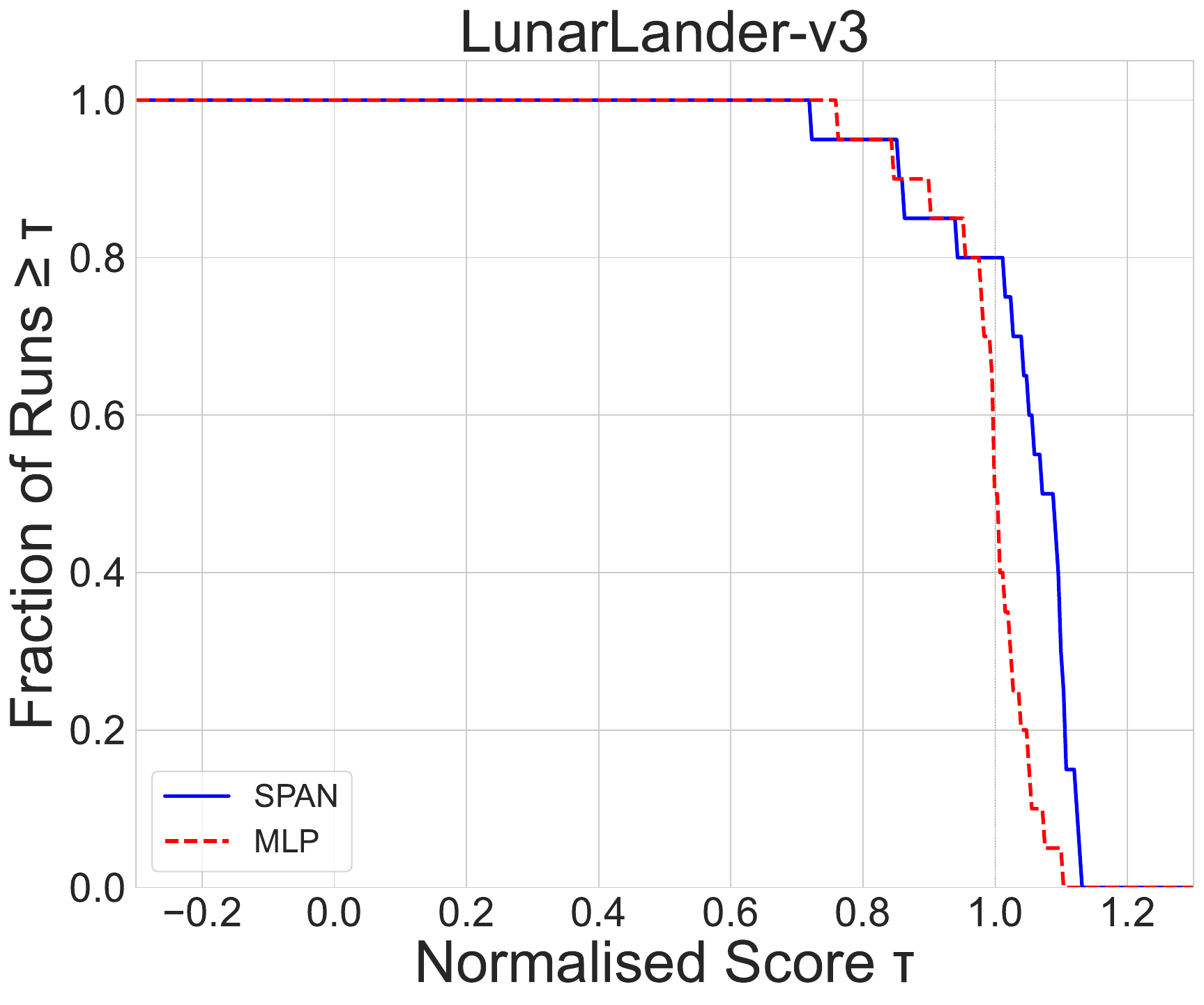}
    \end{subfigure}
    \caption{Per-environment performance profiles for Classic Control tasks. SPAN dominates MLP in all three environments.}
    \label{perf_profiles_classic}
\end{figure}

\begin{figure}[h]
    \centering
    \begin{subfigure}[t]{0.49\textwidth}
        \includegraphics[width=\linewidth]{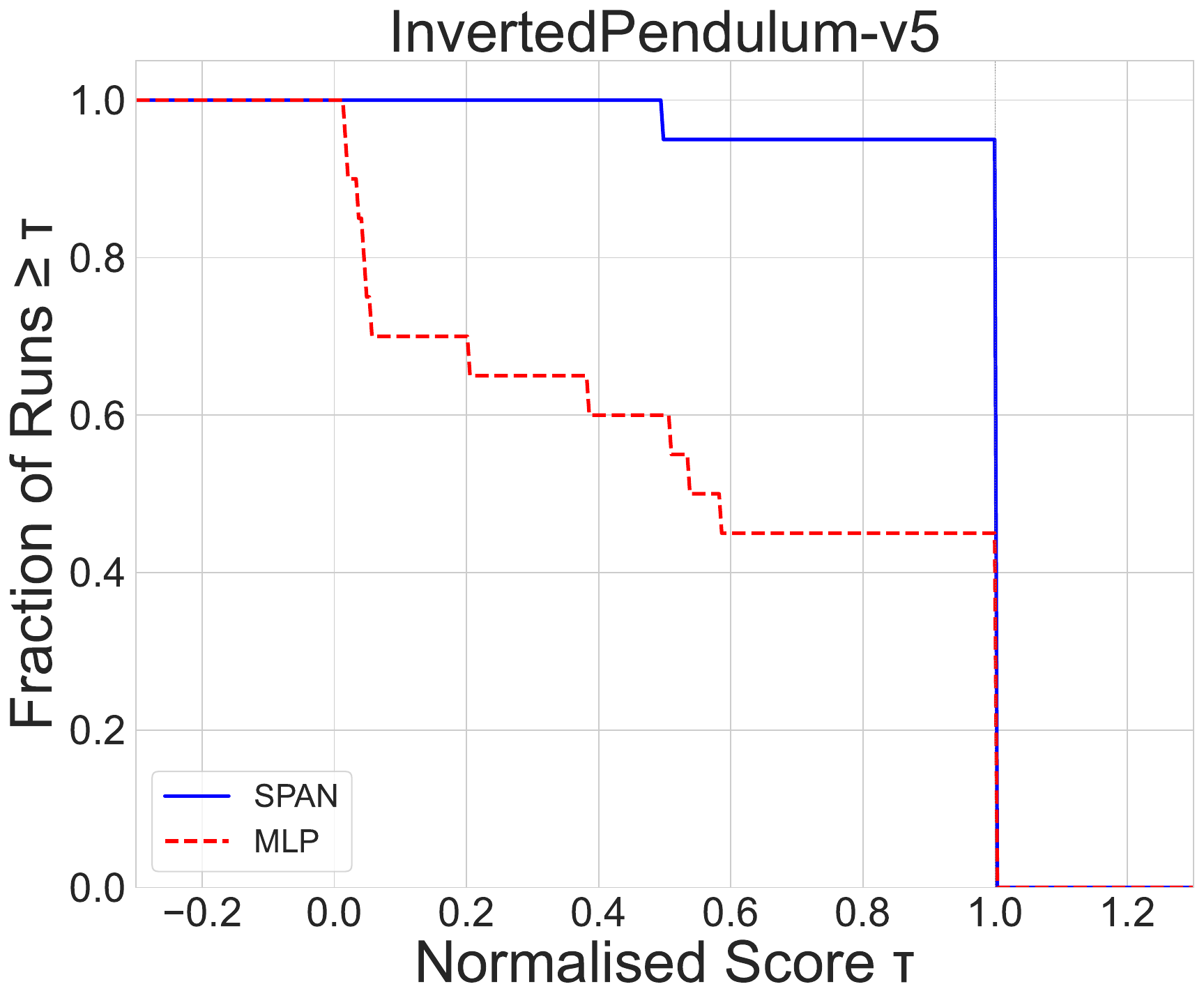}
    \end{subfigure}
    \hfill
    \begin{subfigure}[t]{0.49\textwidth}
        \includegraphics[width=\linewidth]{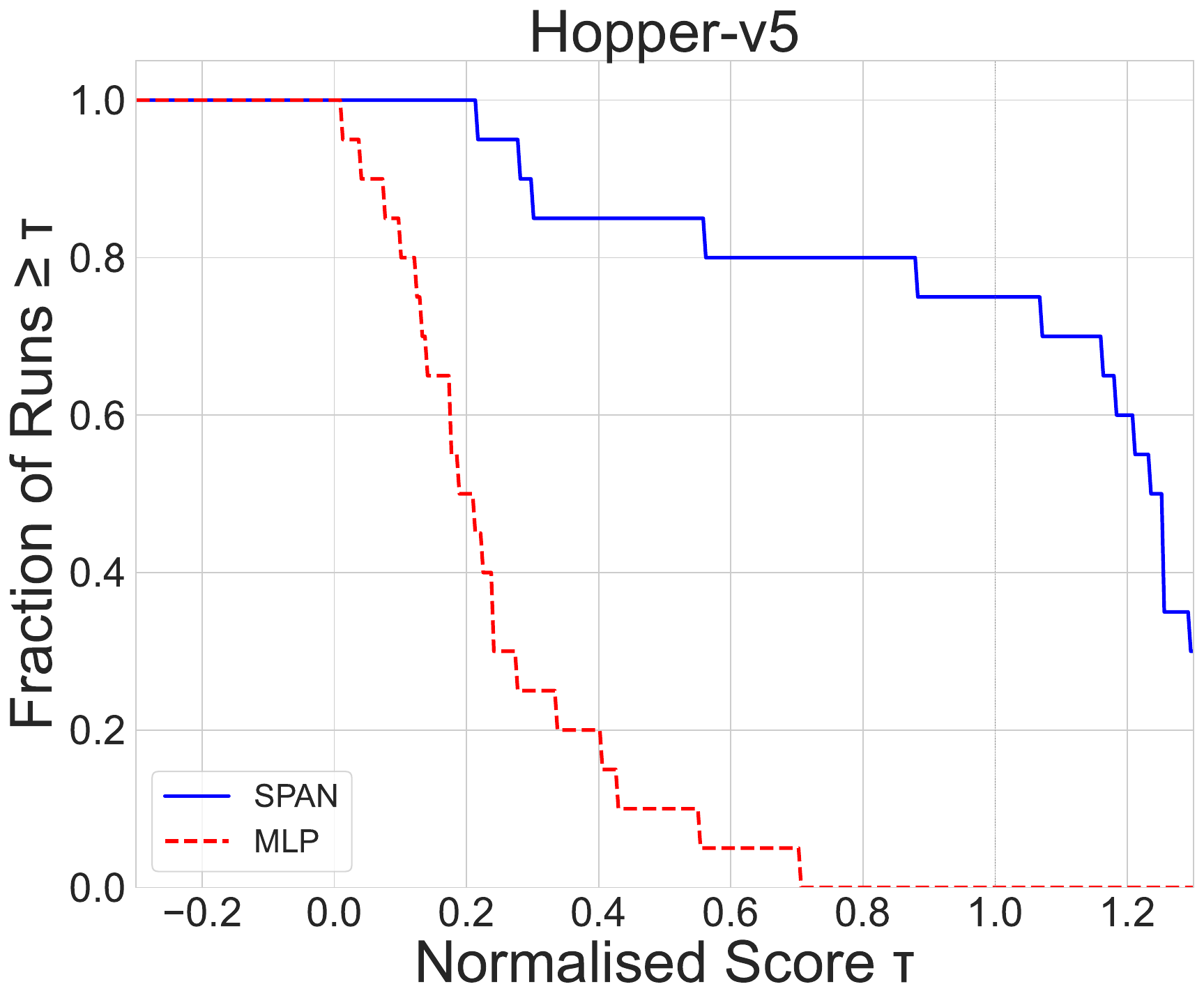}
    \end{subfigure}
    \vskip\baselineskip
    \begin{subfigure}[t]{0.49\textwidth}
        \includegraphics[width=\linewidth]{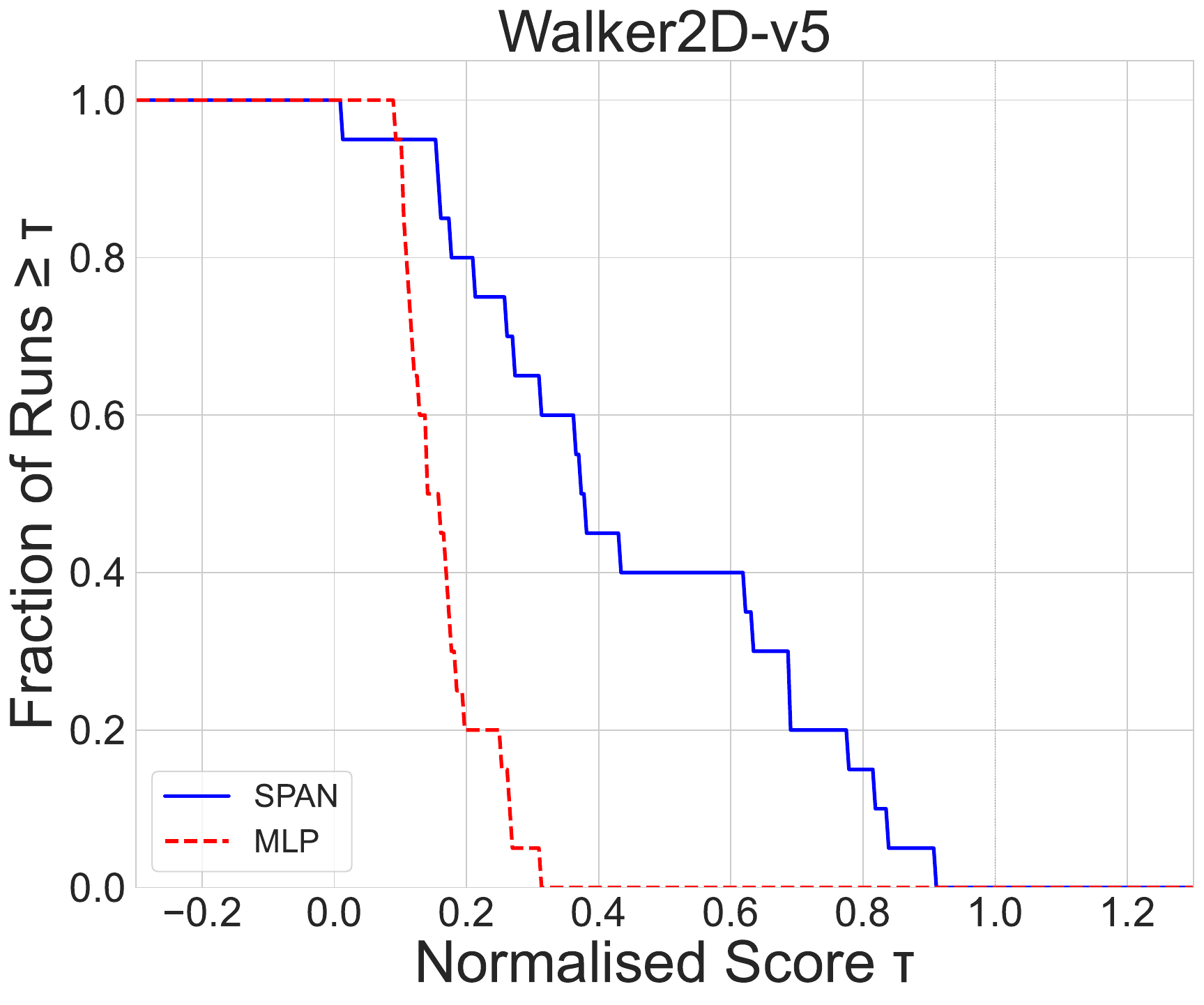}
    \end{subfigure}
    \hfill
    \begin{subfigure}[t]{0.49\textwidth}
        \includegraphics[width=\linewidth]{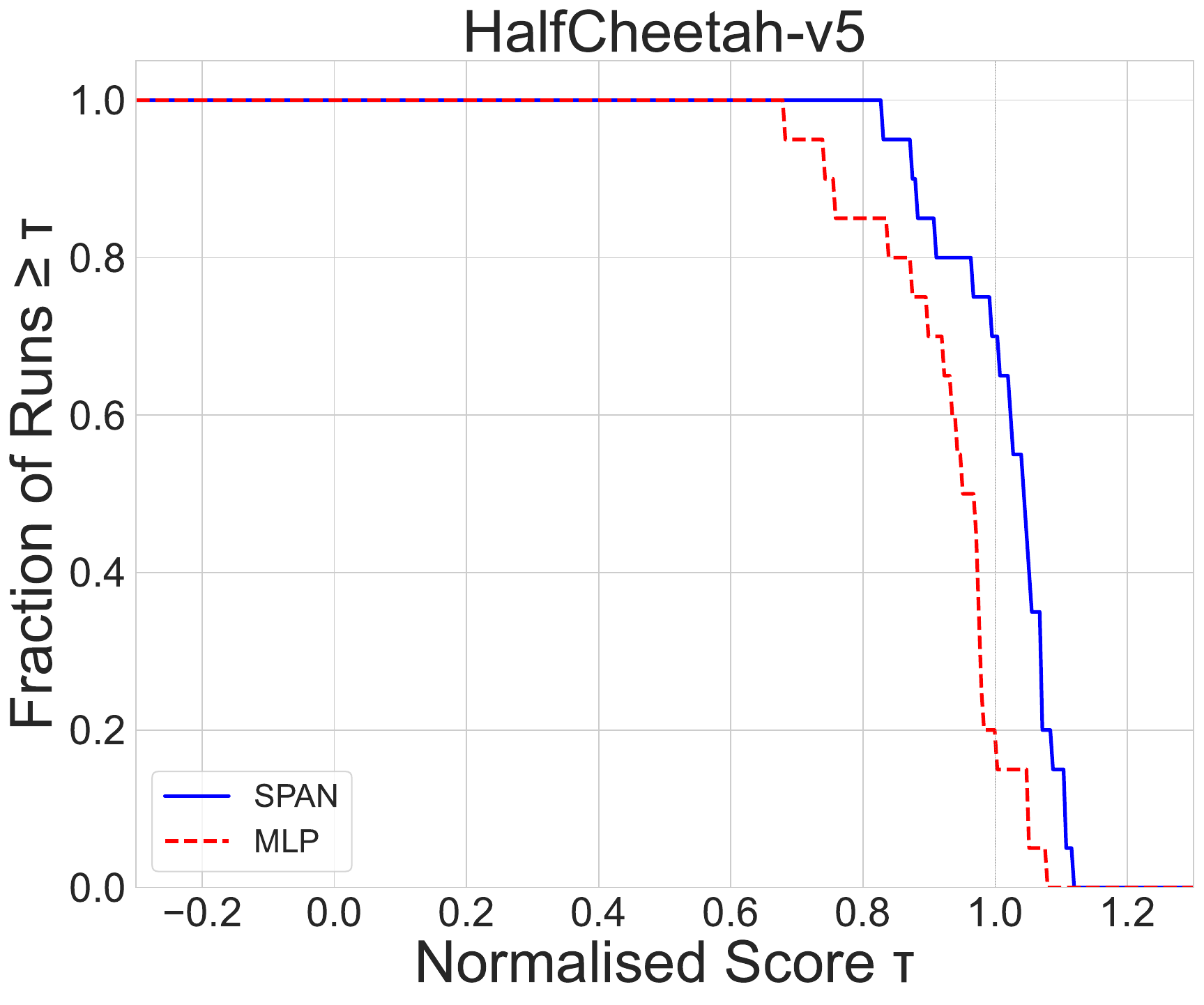}
    \end{subfigure}
    \caption{Per-environment performance profiles for MuJoCo continuous control tasks. SPAN dominates MLP across all four environments.}
    \label{perf_profiles_mujoco}
\end{figure}

\paragraph{Performance profiles.}
Figures ~\ref{perf_profiles_classic} and ~\ref{perf_profiles_mujoco} present per-environment performance profiles for each domain. Within Classic Control (Figure~\ref{perf_profiles_classic}), CartPole shows convergent 
behaviour at moderate thresholds, with the gap emerging only in the tail ($\tau > 0.75$).  Acrobot exhibits a pronounced gap from $\tau = 0$ onward: MLP's fraction of successful runs drops sharply to approximately 70\% at $\tau = 0$ and plateaus, indicating that roughly 30\%  of seeds fail to learn any useful policy, while SPAN maintains $\geq$95\% success across the full mid-range.  LunarLander-v3 shows both agents maintaining a fraction of 1.0 through $\tau \approx 0.7$, with profiles remaining in close agreement until the high-performance tail. Beyond $\tau = 1.0$, SPAN sustains a substantially higher fraction of runs, plateauing near 50\% at $\tau \approx 1.1$ before terminating around $\tau = 1.2$, whereas MLP undergoes a steeper descent and exits the distribution earlier. This indicates that while both architectures reliably solve the task, SPAN produces a heavier tail of high-scoring runs. Note that the fraction of runs $ \geq \tau$ is based solely on final performance, following \citep{agarwal2022}.

\FloatBarrier

Within MuJoCo (Figure~\ref{perf_profiles_mujoco}), InvertedPendulum 
reveals the most striking individual profile: SPAN's curve is flat at 1.0 across the entire range $\tau \in [-0.2, 1.0]$, reflecting its perfect IQM and zero-width CI, while MLP  degrades steadily from $\tau = 0$ and reaches only approximately 45\% of runs at $\tau =  0.5$. Walker2d shows MLP's fraction of successful runs reaching zero at $\tau \approx 0.3$, consistent with the sample efficiency finding that  MLP achieves zero success beyond the 25\% performance threshold, while SPAN's curve descends more gradually, maintaining 40\% of runs above $\tau = 0.5$ and 15\% above $\tau = 0.8$, representing a meaningful capability advantage. Hopper demonstrates the largest capability gap across all environments. MLP's curve collapses to zero at $\tau \approx 0.7$, confirming that no MLP seed achieves expert-level performance within the training budget. SPAN's curve remains above 0.75 through $\tau = 0.8$ and above 0.3 through $\tau = 1.2$, demonstrating that a substantial fraction of seeds not only converge but \emph{exceed} the expert target. On HalfCheetah, both methods achieve high performance on this task, but the profiles diverge in $\tau \in [0.6,\,1.2]$ range. SPAN's curve lies uniformly above MLP's, with non-overlapping IQM CIs confirming that this separation is statistically robust.

\begin{figure}[!hbtp]
    \centering
    \includegraphics[width=0.96\textwidth]{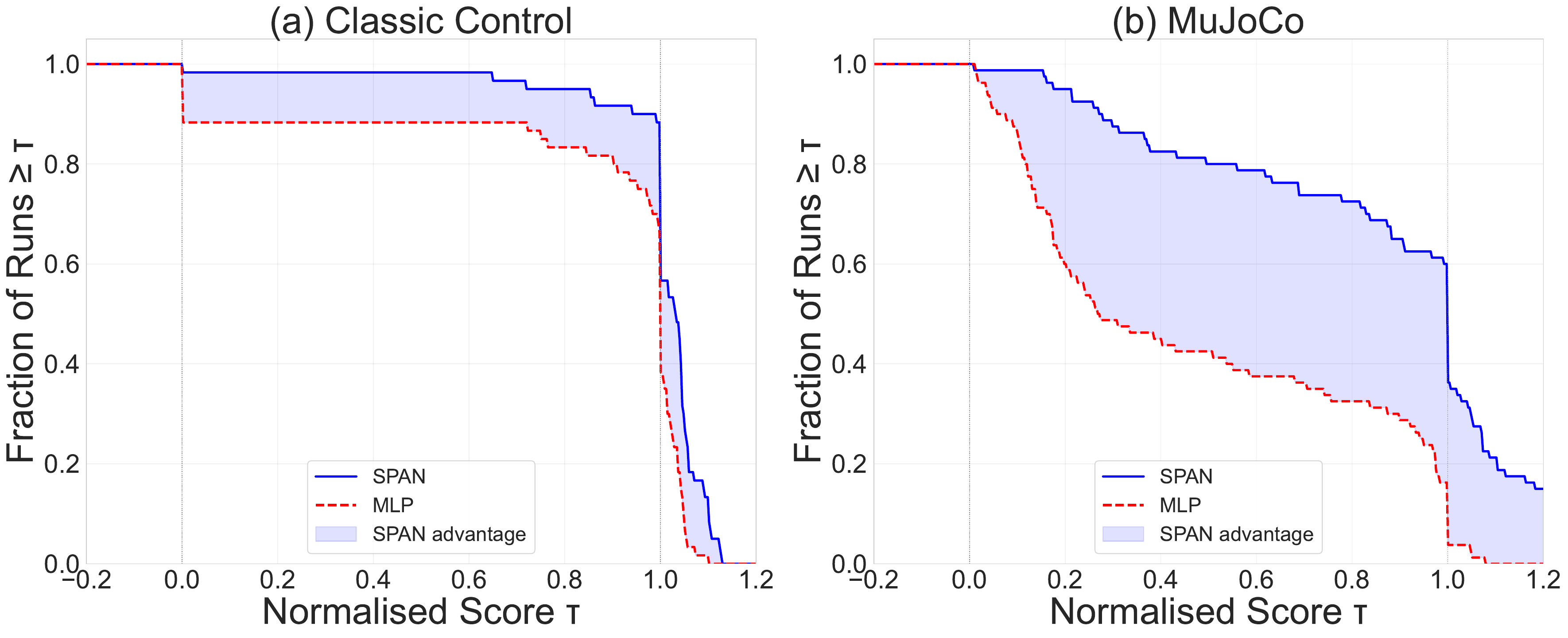}
    \caption{Performance profiles by domain: fraction of runs achieving normalised score $\geq \tau$ across $\tau \in [-0.2,\,1.2]$.}
    \label{fig:perf_profiles}
\end{figure}

The performance profiles in Figure~\ref{fig:perf_profiles} summarizes the qualitatively different patterns by domain. In Classic Control, both architectures achieve high success rates, and SPAN's advantage is concentrated at the right tail of the distribution . In MuJoCo, the gap is substantially larger and emerges from $\tau \approx 0.2$ onward. In both domains, SPAN strictly dominates MLP, confirming that the advantage is not confined to any particular performance band. 

\begin{figure}[hbtp]
    \centering
    \includegraphics[width=0.96\textwidth]{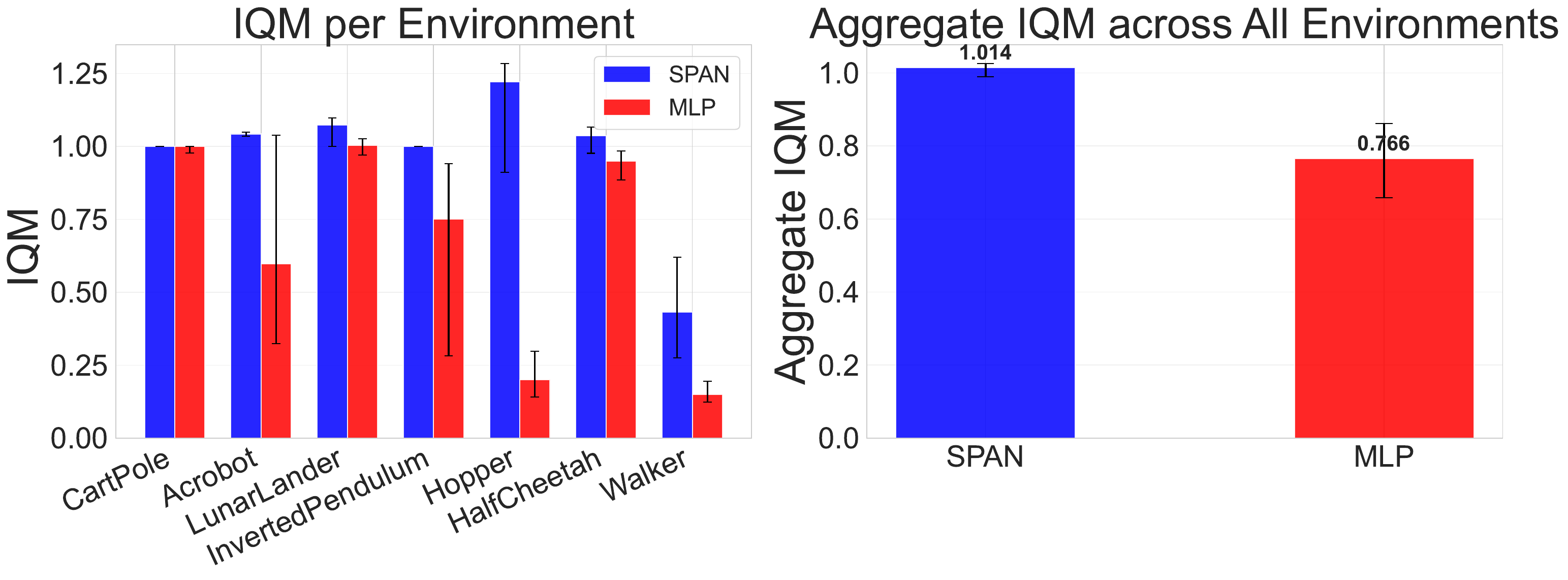}
    \caption{IQM with 95\% bootstrap confidence intervals across 20 seeds.
    \textbf{Left:} Per-environment IQM. 
    \textbf{Right:} Aggregate IQM across all seven environments}
    \label{fig:iqm_comparison}
\end{figure}

\paragraph{Aggregate and per-environment IQM.}
Figure~\ref{fig:iqm_comparison} reports the aggregate IQM across all seven environments and the per-environment breakdown with 95\% bootstrap CIs.
The aggregate IQM is $1.014$ for SPAN versus $0.766$ for the MLP baseline, a gap that the bootstrap CI confirms ( SPAN CI:$[0.989,\,1.026]$, MLP CI:$[0.659,\,0.861]$)  is not attributable to sampling variance.
Per-environment results reveal where the advantage is concentrated.
On Hopper, SPAN achieves an IQM of $1.142$ (CI: $[0.693,\,1.252]$) versus MLP's $0.200$ (CI: $[0.142,\,0.299]$),the MLP fails to learn a stable hopping policy in the majority of seeds.
On  InvertedPendulum, SPAN attains a perfect IQM of $1.000$ with a zero-width CI $[1.000,\,1.000]$, whereas MLP achieves $0.751$ (CI: $[0.280,\,0.941]$), indicating frequent catastrophic failures. On  Walker2d, SPAN's IQM of $0.432$ (CI: $[0.276,\,0.620]$) substantially exceeds MLP's $0.150$ (CI: $[0.124,\,0.195]$), with non-overlapping confidence intervals. On  HalfCheetah, SPAN's CI $[0.977,\,1.067]$ lies entirely above MLP's $[0.885,\,0.985]$, confirming the advantage is robust across all seeds.  CartPole shows a ceiling effect (both IQM = 1.00). On  LunarLander, the IQM gap ($1.073$ vs.\ $1.004$) reflects higher convergence reliability rather than a difference in asymptotic return, consistent with the performance profile in Figure~\ref{fig:perf_profiles}(a).

\begin{table}[H]
\centering
\small
\caption{Per-environment IQM with 95\% bootstrap confidence intervals across 20 seeds. Non-overlapping CIs indicate statistically robust separation.}
\label{tab:iqm_full}
\begin{tabular*}{\columnwidth}{@{\extracolsep{\fill}}lcccc}
\toprule
\textbf{Environment} &
\textbf{SPAN IQM} & \textbf{ SPAN 95\% CI} &
\textbf{MLP IQM}  & \textbf{MLP 95\% CI} \\
\midrule
CartPole         & $1.000$ & $[1.000,\;1.000]$ & $1.000$ & $[0.977,\;1.000]$ \\
Acrobot          & $1.043$ & $[1.035,\;1.048]$ & $0.598$ & $[0.324,\;1.038]$ \\
LunarLander      & $1.073$ & $[1.000,\;1.098]$ & $1.004$ & $[0.970,\;1.026]$ \\
InvertedPendulum & $1.000$ & $[1.000,\;1.000]$ & $0.751$ & $[0.283,\;0.941]$ \\
Hopper           & $1.222$ & $[0.912,\;1.284]$ & $0.200$ & $[0.142,\;0.298]$ \\
HalfCheetah      & $1.037$ & $[0.977,\;1.067]$ & $0.951$ & $[0.885,\;0.985]$ \\
Walker2d         & $0.432$ & $[0.276,\;0.620]$ & $0.150$ & $[0.124,\;0.195]$ \\
\midrule
\textbf{Aggregate} & $\mathbf{1.014}$ & $\mathbf{[0.989,\;1.026]}$ & $\mathbf{0.766}$ & $\mathbf{[0.659,\;0.861]}$ \\
\bottomrule
\end{tabular*}
\end{table}
\paragraph{Significance tests.}
Table~\ref{tab:significance} reports Welch's $t$-tests and Cohen's $d$ effect sizes computed over final-evaluation returns. Four of seven environments reach strong statistical significance ($p < 0.01$), with effect sizes that are large by conventional standards \citep{cohen1988statistical}. On Hopper ($d = 2.709$, $p < 0.0001$) and Walker2d ($d = 1.471$, $p = 0.0001$), the MLP baseline fails entirely at higher performance thresholds, producing near-zero variance that inflates the $t$-statistic; Cohen's $d$ here reflects a genuine capability gap. The two non-significant environments-CartPole ($p = 0.312$) and LunarLander ($p = 0.139$)-are both ceiling cases in which both architectures converge reliably, and the absence of significance reflects saturation rather than equivalence.

\begin{table}[t]
\centering
\small
\caption{Statistical comparison of final-evaluation returns across 20 seeds. Welch's $t$-test with Cohen's $d$ effect size. Significance: $^{*}p<0.05$, $^{**}p<0.01$, $^{***}p<0.001$, ns = not significant.}
\label{tab:significance}
\begin{tabular*}{\columnwidth}{@{\extracolsep{\fill}}lccccc}
\toprule
\textbf{Environment} &
\textbf{SPAN} ($\mu \pm \sigma$) &
\textbf{MLP} ($\mu \pm \sigma$) &
$t$ &
$p$ &
$d$ \\
\midrule
CartPole          & $491 \pm 39$    & $464 \pm 112$  & $1.033$ & $0.312$\;\text{ns} & $0.327$\;\text{ns} \\
Acrobot           & $-104 \pm 93$   & $-216 \pm 193$ & $2.322$ & $0.028^{*}$        & $0.734^{*}$ \\
LunarLander       & $218 \pm 55$    & $195 \pm 39$   & $1.516$ & $0.139$\;\text{ns} & $0.480$\;\text{ns} \\
InvertedPendulum  & $975 \pm 113$   & $572 \pm 432$  & $4.036$ & $0.001^{***}$      & $1.276^{***}$ \\
Hopper            & $2647 \pm 971$  & $611 \pm 431$  & $8.567$ & $<\!0.00^{***}$   & $2.709^{***}$ \\
HalfCheetah       & $6099 \pm 527$  & $5544 \pm 659$ & $2.939$ & $0.006^{**}$       & $0.929^{**}$ \\
Walker2d          & $1589 \pm 942$  & $582 \pm 222$  & $4.653$ & $<\!0.001^{***}$   & $1.471^{***}$ \\
\bottomrule
\end{tabular*}
\end{table}

% ============================================================
% Section 5.3 — Computational Cost Analysis
% ============================================================

\subsection{Computational Cost Analysis}
\label{sec:compute}

Sample efficiency advantages are practically meaningful only if they translate to
lower total computational cost. We assess this through three complementary metrics:
steps-per-second (SPS) throughput, raw wall-clock time to reach the 100\% performance threshold, and \emph{expected training cost}-defined as wall-clock time to threshold divided by the empirical success rate-which accounts for the probability that a given training run will converge at all. Peak training memory is identical between SPAN and MLP at matched parameter budgets, as model parameters, optimiser states, and replay buffers are all equal; we therefore
focus on time-based costs.

\paragraph{Per-step throughput.}
KHRONOS evaluations introduce a per-step overhead. Measured SPS(Steps Per Second) ratios (MLP/SPAN) range from $1.2$ - $2.1$ (Table~\ref{tab:compute}), meaning SPANS takes about $1.2$ - $2.1$ more times completing one single step in the RL environment. 

\paragraph{Wall-clock time to threshold.}
Despite slower per-step throughput, SPAN's superior sample efficiency yields shorter or comparable raw wall-clock times to reach the 100\% performance threshold in most environments. SPAN converges strictly faster on CartPole ($13.7$ vs.\ $14.5$ min), Acrobot ($12.3$ vs.\ $14.0$ min), InvertedPendulum ($36.8$ vs.\ $53.9$ min), and Walker2d ($631.5$ vs.\ $388.2$ min, noting that MLP achieves only 20\% success at the 25\% threshold and zero thereafter).
On LunarLander and HalfCheetah, MLP holds a slight raw-time edge at the 100\%
threshold ($56.3$ vs.\ $60.5$ min; $627.6$ vs.\ $773.5$ min respectively).
On Hopper, SPAN converges in $315$ min while MLP fails to converge in any seed.

\paragraph{Expected training cost.}
The most practically relevant metric is the expected cost to produce \emph{one successfully trained agent}, which penalises architectures whose lower per-run time is offset by high failure rates:
\begin{equation}
    \mathcal{C}_{\text{expected}} = \frac{T_{\text{threshold}}}{\text{SR}},
    \label{eq:expected_cost}
\end{equation}
where $T_{\text{threshold}}$ is the wall-clock time to the performance threshold and $\text{SR}$ is the empirical success rate across seeds. This metric reveals the decisive advantage of SPAN across all environments
(Table~\ref{tab:compute}). On HalfCheetah, the MLP's expected cost is $6{,}276$ min versus SPAN's $1{,}289$ min-a $4.87\times$ increase-because MLP's 10\% success rate at the 100\% threshold renders its faster per-run time irrelevant.
On LunarLander, the MLP's raw-time advantage ($56.3$ vs.\ $60.5$ min) reverses
completely: its 60\% success rate yields an expected cost of $93.8$ min compared to
SPAN's $67.2$ min ($1.40\times$ more expensive). On Hopper, MLP achieves only a 10\% success rate at the 50\% threshold and zero beyond it, while on Walker2d, it manages just 20\% success at the lowest threshold (25\%) and zero thereafter, rendering its expected cost infinite after those threshold. Overall, accounting for convergence reliability, training an MLP is $1.34\times$ to $6.29\times$ more computationally expensive than SPAN across environments where both methods eventually converge, and infinitely more expensive on Hopper and Walker2d at higher performance thresholds.

\begin{table*}[t]
\centering
\small
\caption{Computational cost analysis at the 100\% performance threshold.
SPS = steps per second (higher is faster per step).
SPS Ratio = MLP/SPAN (higher means SPAN is slower per step).
$T_{100\%}$ = wall-clock time to threshold.
SR = success rate at threshold.
$\mathcal{C}_{\text{expected}}$ = expected cost (Eq.~\ref{eq:expected_cost}).
Cost Ratio = MLP/SPAN expected cost (higher means MLP is more expensive overall).
$^\dagger$Hopper values are reported at 50\% threshold, the highest at which MLP achieves any success .
$^\ddagger$Walker2d values reported at the 25\% threshold, the highest at which MLP achieves any success.}
\label{tab:compute}
\begin{tabular*}{\textwidth}{@{\extracolsep{\fill}}lccccccccccc}
\toprule
& \multicolumn{2}{c}{\textbf{SPS}} & & \multicolumn{2}{c}{$T_{100\%}$ \textbf{(min)}} & \multicolumn{2}{c}{\textbf{SR at} $100\%$} & \multicolumn{2}{c}{$\mathcal{C}_{\text{expected}}$ \textbf{(min)}} & \\
\cmidrule(lr){2-3}\cmidrule(lr){5-6}\cmidrule(lr){7-8}\cmidrule(lr){9-10}
\textbf{Environment} & SPAN & MLP & \textbf{SPS Ratio} & SPAN & MLP & SPAN & MLP & SPAN & MLP & \textbf{Cost Ratio} \\
\midrule
CartPole           & 245 & 346 & $1.41\times$ & 13.7  & 14.5  & 95\% & 75\%  & 14.4   & 19.3     & $1.34\times$ \\
Acrobot            & 197 & 271 & $1.38\times$ & 12.3  & 14.0  & 90\% & 60\%  & 13.7   & 23.3     & $1.70\times$ \\
LunarLander        & 170 & 240 & $1.41\times$ & 60.5  & 56.3  & 90\% & 60\%  & 67.2   & 93.8     & $1.40\times$ \\
InvertedPendulum   & 45  & 55  & $1.22\times$ & 36.8  & 53.9  & 100\%& 100\% & 36.8   & 53.9     & $1.46\times$ \\
HalfCheetah        & 15  & 24  & $1.60\times$ & 773.5 & 627.6 & 60\% & 10\%  & 1289.2 & 6276.0   & $4.87\times$ \\
Hopper$^\dagger$   & 26  & 56  & $2.15\times$ & 304.5 & 212.8 & 90\% & 10\%  & 338.3  & 2128     & $6.29\times$ \\
Walker2d$^\ddagger$& 14  & 25  & $1.79\times$ & 631.5 & 388.2 & 65\% & 20\%  & 971.5  & 1941.0   & $2.00\times$ \\
\bottomrule
\end{tabular*}
\end{table*}

\FloatBarrier

\subsection{Offline Reinforcement Learning Results}
To assess SPAN's effectiveness beyond online learning, SPAN is also evaluated on offline RL using Implicit Q-Learning (IQL) \citep{kostrikov2021iql} on the D4RL \citep{fu2021d4rl} Adroit Hand Manipulation benchmark from Minari \citep{minari}. 

Table~\ref{tab:d4rl_normalized} presents normalized performance scores across four dexterous manipulation tasks (Door, Hammer, Pen, Relocate) with three dataset qualities: Expert (near-optimal demonstrations), Cloned (behavioral cloning data), and Human (suboptimal human teleoperations).

\FloatBarrier
\begin{table}[t]
\centering
\caption{Normalized Offline RL Performance on Minari Adroit (v2). Scores are normalized using standard D4RL reference constants. SPAN outperforms the MLP baseline on all Expert datasets.}
\label{tab:d4rl_normalized}
\small
\begin{tabular}{llcc}
\toprule
\textbf{Task} & \textbf{Dataset} & \textbf{SPAN (Ours)} & \textbf{MLP (Baseline)} \\
\midrule
\multirow{3}{*}{Door} & Expert & \textbf{8.0} $\pm$ 11.6 & 0.4 $\pm$ 0.0 \\
 & Cloned & \textbf{0.6} $\pm$ 0.1 & \textbf{0.6} $\pm$ 0.1 \\
 & Human & 0.5 $\pm$ 0.1 & \textbf{1.2} $\pm$ 1.3 \\
\midrule
\multirow{3}{*}{Hammer} & Expert & \textbf{44.1} $\pm$ 38.3 & 19.3 $\pm$ 23.4 \\
 & Cloned & \textbf{0.3} $\pm$ 0.0 & \textbf{0.3} $\pm$ 0.0 \\
 & Human & 0.5 $\pm$ 0.3 & \textbf{0.9} $\pm$ 0.6 \\
\midrule
\multirow{3}{*}{Pen} & Expert & \textbf{124.1} $\pm$ 19.8 & 119.6 $\pm$ 19.7 \\
 & Cloned & \textbf{8.8} $\pm$ 19.2 & 6.7 $\pm$ 13.3 \\
 & Human & \textbf{21.1} $\pm$ 31.8 & -10.0 $\pm$ 6.1 \\
\midrule
\multirow{3}{*}{Relocate} & Expert & \textbf{17.3} $\pm$ 8.6 & 5.2 $\pm$ 3.0 \\
 & Cloned & \textbf{0.4} $\pm$ 0.1 & \textbf{0.4} $\pm$ 0.0 \\
 & Human & \textbf{0.3} $\pm$ 0.0 & \textbf{0.3} $\pm$ 0.0 \\
\bottomrule
\end{tabular}
\end{table}
\vspace{1em}

 SPAN consistently outperforms MLP baselines on Expert datasets, achieving substantial improvements on challenging tasks: Door (8.0 vs 0.4, 20× improvement), Hammer (44.1 vs 19.3, 2.3× improvement), and Relocate (17.3 vs 5.2, 3.3× improvement). On Pen, both methods achieve strong expert-level performance (SPAN: 124.1, MLP: 119.6), with SPAN showing marginally higher scores. Performance on lower quality datasets (Cloned, Human) reveals more mixed results. SPAN maintains advantages on Pen across all dataset types (8.8 vs 6.7 on Cloned, 21.1 vs -10.0 on Human), suggesting robustness to demonstration quality. However, on Door, Hammer human datasets, MLP baselines show slight edges. This asymmetry reflects the interaction between SPAN's smoothness bias and dataset characteristics: expert demonstrations exhibit smooth, consistent policies that align with KHRONOS approximation, while human datasets contain erratic behaviors and sharp transitions that challenge smooth function approximation. These results demonstrate SPAN's advantages extend beyond online learning to offline RL, particularly when learning from high quality demonstrations. The dramatic improvements on expert datasets (2-20× on three of four tasks) suggest SPAN is especially valuable where smooth behaviors dominate the dataset.

\subsection{Ablation Studies}
To characterize the sensitivity of SPAN to its structural hyperparameters, grid searches are conducted . Three key architectural parameters: the tensor rank ($M$, corresponding to nmodes), the grid resolution ($N$, corresponding to nelems), and the polynomial degree ($k$) are varied. Acrobot, LunarLander, Hopper and HalfCheetah are presented here as a representative results, and  ablation results for other environments are provided in Appendix~\ref{app:ablation}. Each panel shows the effect of varying a single architectural hyperparameter while keeping all other settings fixed.The distributions are shown as box plots over multiple independent training runs. In each box plot, the central line denotes the median final evaluation return, the box spans the interquartile range (25th-75th percentiles), and the whiskers indicate the full range of observed values.

Figure~\ref{fig:acrobot_ablation} illustrates the hyperparameter sensitivity for Acrobot. Performance remains remarkably stable across all configurations. Increasing the rank from $M=2$ to $M=10$ or resolution from $N=2$ to $N=5$ yields negligible changes in final reward. This suggests that for lower dimensional control tasks, the low rank of SPAN is sufficient to capture the optimal policy even at minimal capacity.

\begin{figure*}[t]
\centering
\begin{subfigure}[b]{0.96\textwidth}
    \includegraphics[width=\textwidth]{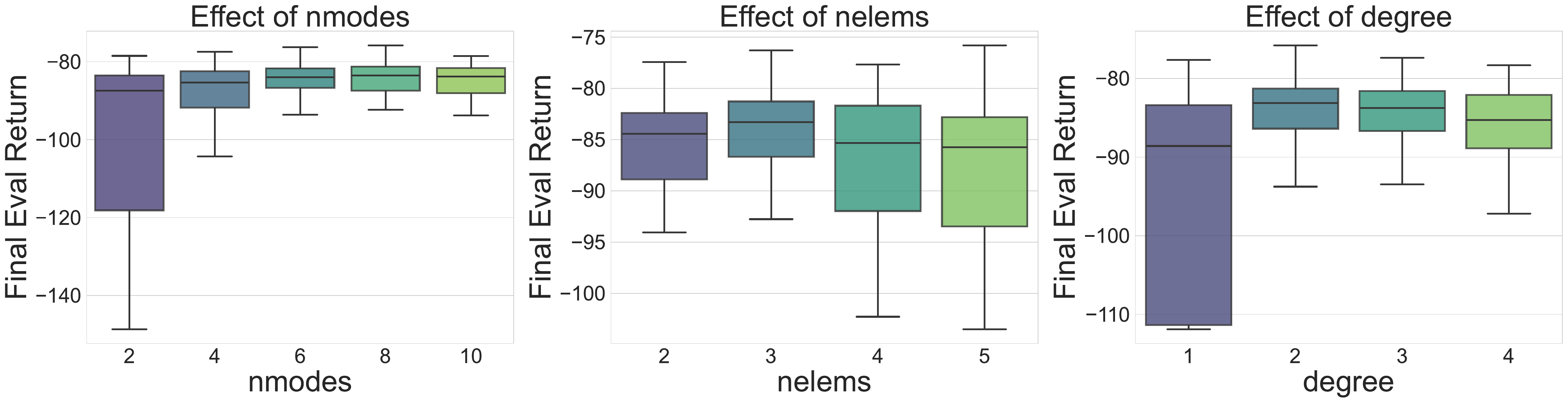}
    %\caption{CartPole}
    \label{fig:acrobot_ablation}
\end{subfigure}
\caption{Ablation study on Acrobot-v1 showing performance vs. nmode, nelem, and degree.} 
\label{fig:acrobot_ablation}
\end{figure*}

\begin{figure*}[t]
\centering
\begin{subfigure}[b]{0.96\textwidth}
    \includegraphics[width=\textwidth]{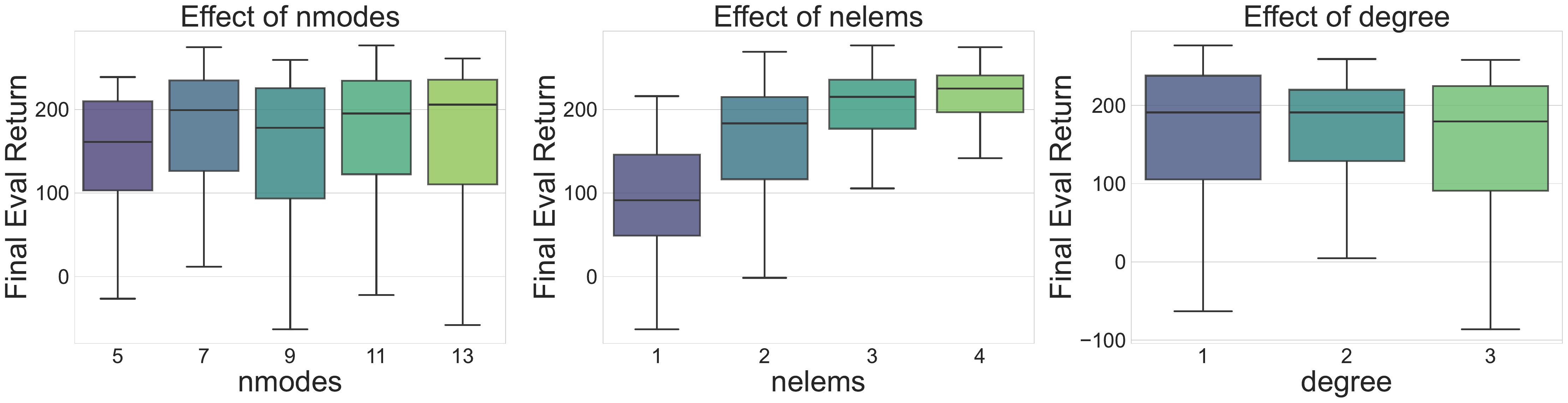}
    %\caption{CartPole}
    \label{fig:LunarLander_ablation}
\end{subfigure}
\caption{Ablation study on LunarLander-v3 showing performance vs. nmodes, nelems, and degree.}
\label{fig:LunarLander_ablation}
\end{figure*}

Figure \ref{fig:LunarLander_ablation} exhibits sensitivity patterns of LunarLander environment. Performance improves progressively with nmodes, increasing from $\sim$150 return at nmodes=5 to $\sim$200 at nmodes=13, indicating that this moderately complex task benefits from additional representational modes. The nelems parameter displays a clear trend: nelems=1 yields poor performance ($\sim$100 return) while nelems $\geq$ 4 achieves stable performance ($\sim$210 return). The degree parameter remains relatively invariant across [1,3], suggesting polynomial order has minimal impact once other capacity requirements are satisfied.

\begin{figure}[h!]
    \centering
    \includegraphics[width=0.96\textwidth]{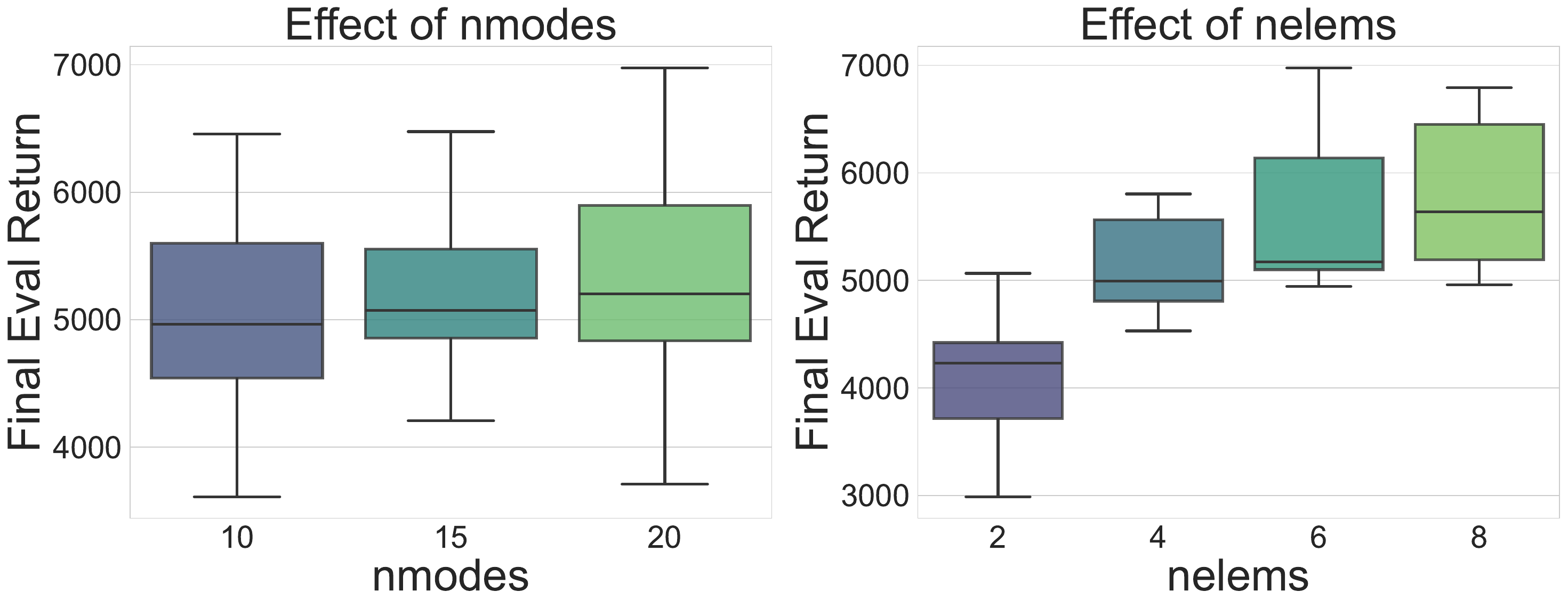}
    \caption{Ablation study on HalfCheetah-v5 showing performance vs. nmode, nelem.}
    \label{fig:halfcheetah_ablation}
\end{figure}

For the 17-dimensional HalfCheetah task, optimized SPAN implementation is used, which fixes the spline degree to quadratic ($k=2$) to focus the search space on rank and resolution. As shown in Figure~\ref{fig:halfcheetah_ablation}, this environment exhibits a distinct scaling law. Unlike Acrobot, HalfCheetah benefits significantly from increased resolution, with returns improving consistently as grid density increases from $N=2$ to $N=8$. Similarly, increasing tensor rank from $M=10$ to $M=20$ yields a positive trend in asymptotic performance. 

\textbf{Hopper-v5} demonstrates substantial capacity requirements with high performance variance. Increasing nmode from 6 to 10 shows clear improvements in median performance alongside marked reduction in failure cases. The nelem parameter exhibits particularly strong effects: nelem=4 produces highly variable outcomes (500-2500 return range), while nelem=8 achieves both superior and more consistent results (median $\sim$3000 return with substantially tighter distribution), underscoring the importance of adequate capacity for complex continuous control.

\begin{figure*}[!htbp]
\centering
\begin{subfigure}[b]{0.96\textwidth}
    \includegraphics[width=\textwidth]{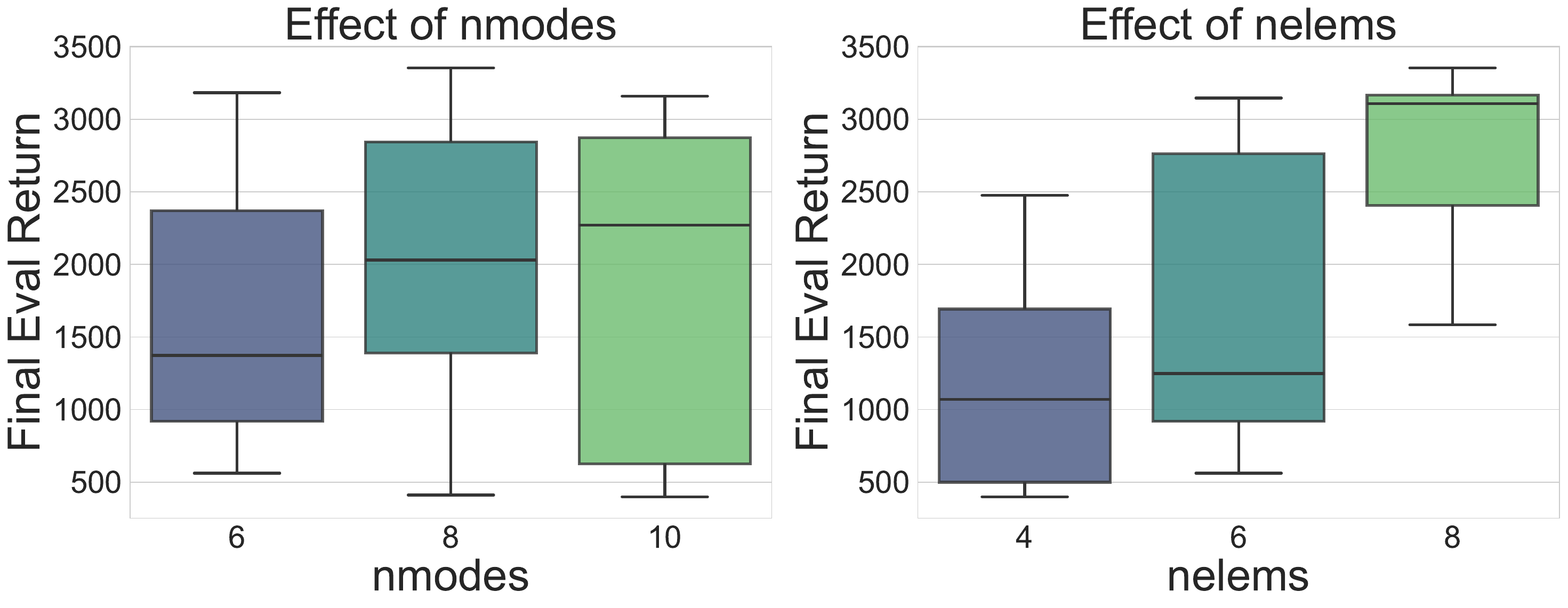}
    %\caption{CartPole}
    \label{fig:Hopper_ablation}
\end{subfigure}
\caption{Ablation study Hopper showing performance vs. nmode and nelem}
\label{fig:Hopper_ablation}
\end{figure*}

\section{Application: Data Center HVAC Control}
\label{sec:application}

HVAC systems account for over 50\% of building energy demand in developed countries, making them a high-impact target for intelligent control  \citep{campoy2025sinergym}. Within this broader context, data centres constitute a particularly high-impact subclass: global data centre electricity consumption reached 415\,TWh in 2024 approximately 1.5\% of world electricity use and is projected to double to 945\,TWh by 2030, driven primarily by AI workload growth \citep{iea2025energyai}. Cooling infrastructure accounts for 7-30\% of total data centre energy demand \citep{iea2025energyai}, making HVAC control a direct lever on this trajectory. Data centre HVAC control poses a resource-constrained reinforcement learning problem in a precise sense. Every control decision raising or lowering a cooling setpoint incurs an immediate, non-recoverable 
energy expenditure while simultaneously determining whether zone temperatures remain within the comfort band over the subsequent timesteps. Energy consumed cannot be reclaimed within an episode, and comfort violations accumulate irreversibly: a zone that overheats during a transition month cannot be retroactively corrected. 

Two properties of this domain make it a natural stress test for SPAN specifically. First, the relationship between cooling setpoint and zone temperature response is locally smooth - governed by the building's thermal mass and thermodynamic exchange with the outdoor environment.  A controller that approximates this smooth structure accurately translates into energy efficiency. Second, real data centre operators cannot afford thousands of trial episodes before deploying a controller: each episode represents a full year of building simulation, and the operational cost of suboptimal 
HVAC control during policy search measured in both energy expenditure and comfort violations accumulates with every interaction required to learn. Sample efficiency is therefore not an abstract benchmark property but an operational requirement: a method that learns a high-quality setpoint policy in fewer episodes directly reduces the energy wasted during training. Together, these two properties smooth thermodynamic structure and a limited interaction budget define a regime in which 
SPAN's inductive bias is not merely convenient but functionally consequential. The energy and comfort results reported below constitute a direct empirical test of whether this architectural alignment translates into measurable operational gains.

\paragraph{Environment.} The evaluation is conducted using Sinergym 
\citep{campoy2025sinergym}, an open-source virtual testbed for reinforcement 
learning-based building energy optimisation built on the EnergyPlus simulation engine and the Gymnasium interface. Sinergym provides standardised environments, customisable reward functions, weather variability, and logging infrastructure that enable reproducible comparison across controllers. The specific environment used is \texttt{Eplus-datacenter\_cw-mixed-continuous-v1}, which simulates the \texttt{2ZoneDataCenterHVAC} building mode a 491.3\,m$^2$ two-zone facility with air economisers, evaporative coolers, DX and chilled water cooling coils, and VAV units driven by New York TMY3 weather data (USA NY New York J.F.\ Kennedy). The observation space is 41-dimensional, encompassing outdoor meteorological variables (dry-bulb temperature, 
relative humidity, wind speed, direct and diffuse solar radiation), zone air temperatures and humidity, HVAC setpoints, and total facility electricity demand. The continuous action space controls two variables: the cooling setpoint temperature and the chilled water supply temperature.

\paragraph{Reward function.} The reward at each timestep linearly combines HVAC electricity 
demand and a thermal comfort penalty:
\begin{equation}
    r_t = -\omega \lambda_P P_t - (1 - \omega) \lambda_T \left(|T_t - T_{up}| + |T_t - T_{low}|\right),
\end{equation}
where $P_t$ represents the HVAC electrical power consumption (W), and $T_t$ is the current indoor air temperature ($^\circ$C) for the west and east zones. The imposed comfort range limits, $T_{low} = 18.0^\circ$C and $T_{up} = 27.0^\circ$C. The baseline configuration uses an energy weight of $\omega = 0.5$, an energy scaling constant of $\lambda_P = 0.00005$, and a comfort scaling constant of $\lambda_T = 1.0$. The thermal penalty is exactly $0$ if the zone temperature $T_t$ remains within the imposed comfort range.

\begin{figure}[t]
    \centering
    \includegraphics[width=0.5\linewidth]{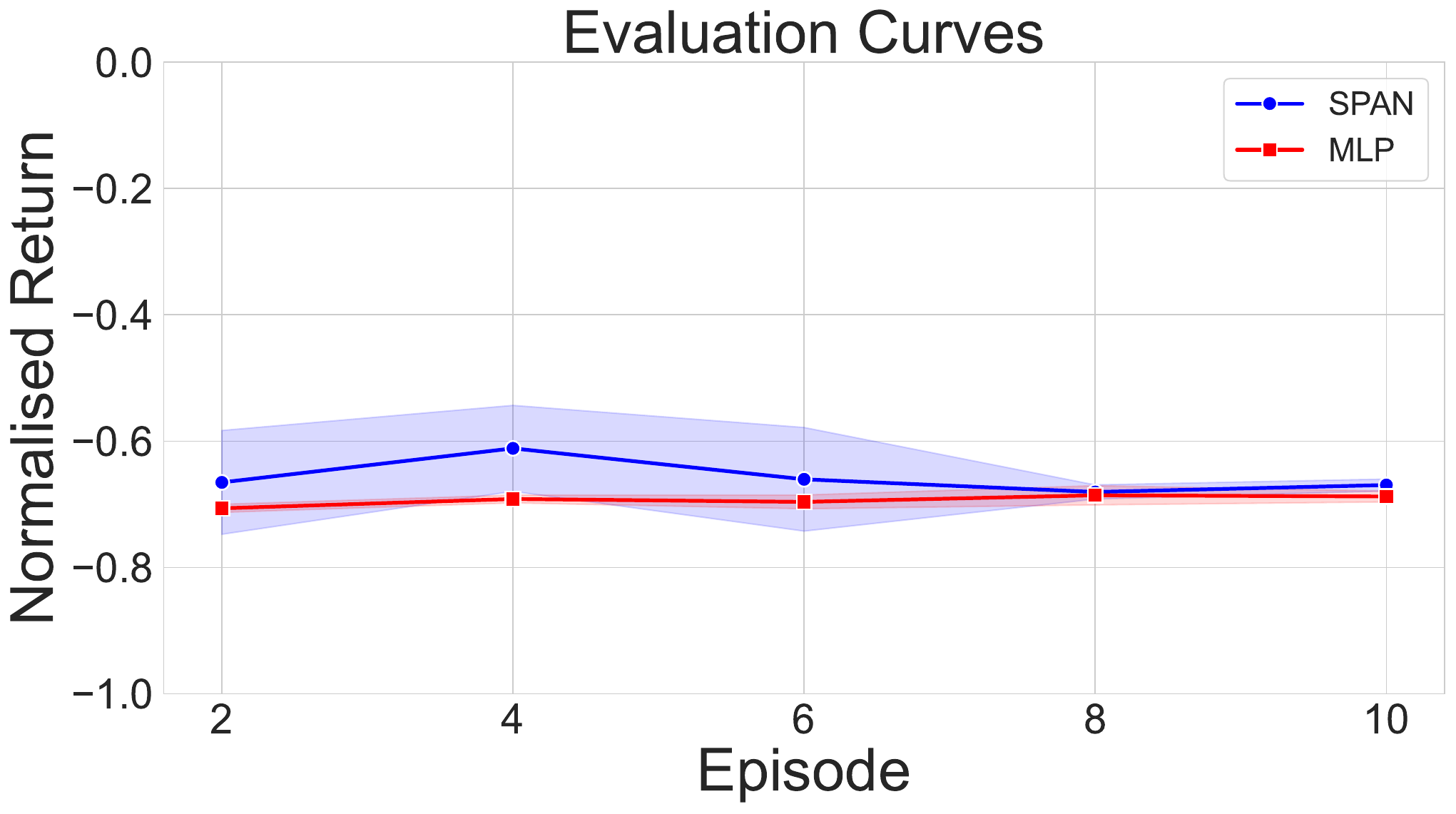}
    \caption{Normalised evaluation return per training episode for SAC-SPAN and SAC-MLP 
    on the Sinergym datacenter environment, averaged over 5 seeds. SPAN achieves consistently 
    higher return throughout training.}
    \label{fig:sinergym_eval}
\end{figure}

\paragraph{Results.}
Figure~\ref{fig:sinergym_eval} reports the normalised evaluation return across training  episodes averaged over five seeds. Because the reward function penalises both HVAC electricity consumption and thermal comfort deviation simultaneously a higher (less negative) normalised return directly reflects a controller that reduces energy use while maintaining zone temperatures within the comfort band $[18, 27]^\circ$C. SPAN achieves a consistently higher normalised 
return throughout training.

\begin{figure*}[t]
    \centering
    \begin{subfigure}[b]{0.48\textwidth}
        \includegraphics[width=\linewidth]{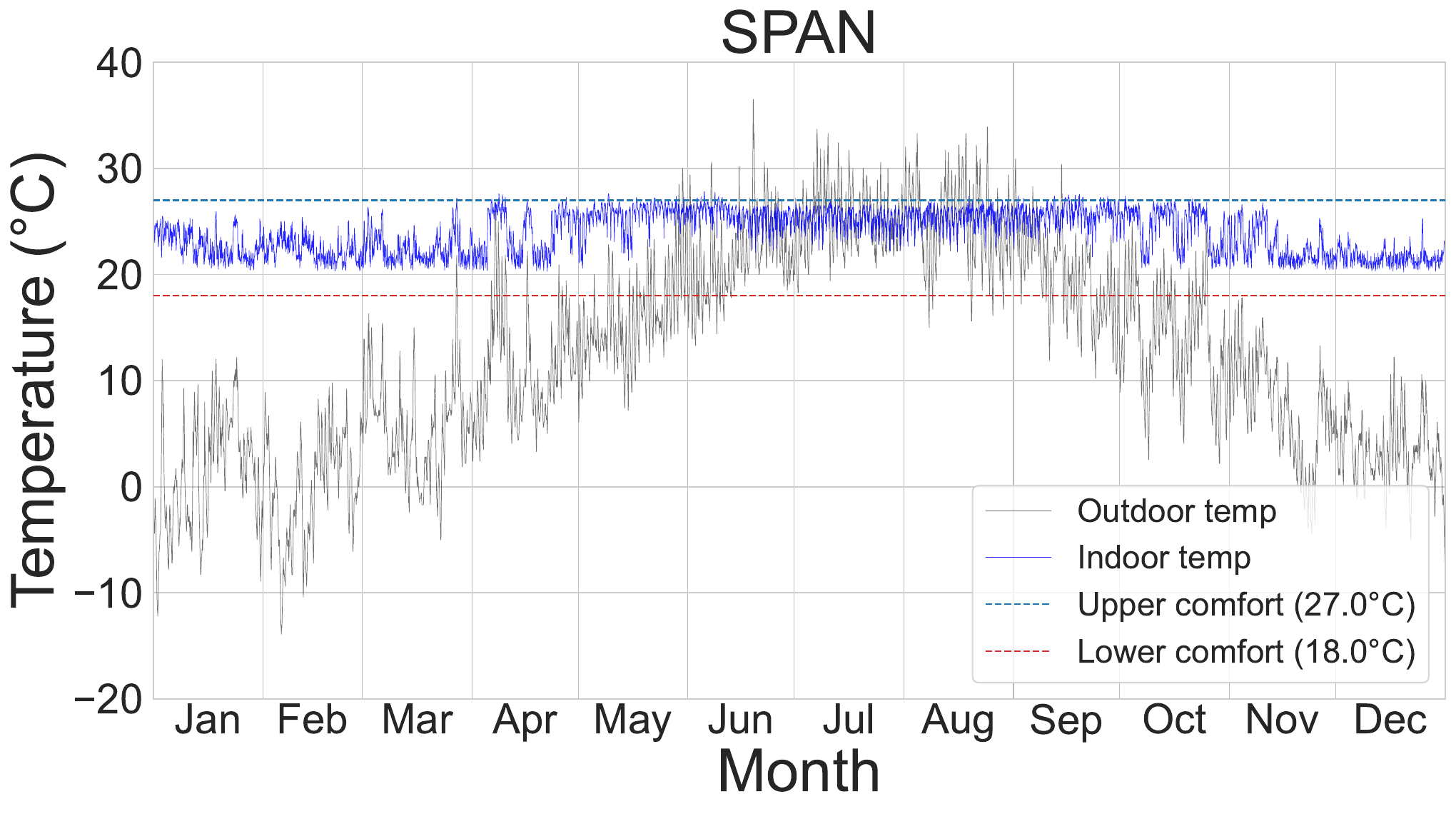}
        %\caption{SAC-SPAN}
    \end{subfigure}
    \hfill
    \begin{subfigure}[b]{0.48\textwidth}
        \includegraphics[width=\linewidth]{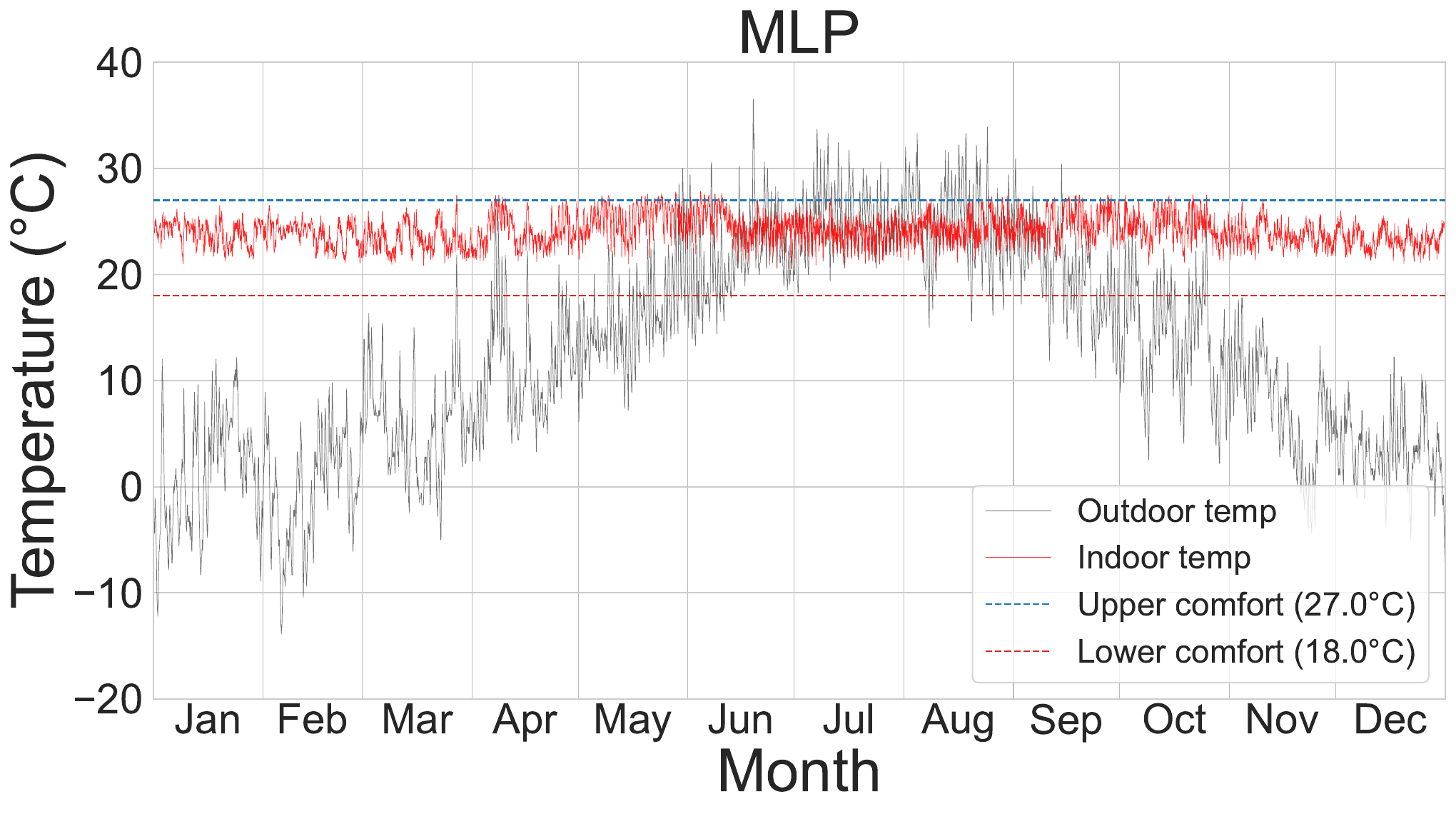}
        %\caption{SAC-MLP}
    \end{subfigure}
    \caption{Indoor air temperature (coloured) and outdoor dry-bulb temperature (grey) 
    over the full evaluation year, sampled at every four hour.}
    \label{fig:sinergym_temp}
\end{figure*}

Figure~\ref{fig:sinergym_temp} shows the indoor air temperature maintained by each controller over the full evaluation year, plotted alongside the outdoor dry-bulb temperature and the ASHRAE comfort bounds. Outdoor temperatures range from below $-10^\circ$C in winter to above $35^\circ$C in summer, yet both controllers hold indoor conditions within the comfort band across all season a direct consequence of learning to modulate cooling setpoints in response to external thermal load rather than applying static schedules. Visually, the MLP controller (red) more frequently approaches the upper comfort boundary of $27^\circ$C during the transition months of May, June, and September, while SPAN (blue) tracks closer to the center of the band during the same periods.

\begin{figure}[t]
        \includegraphics[width=\linewidth]{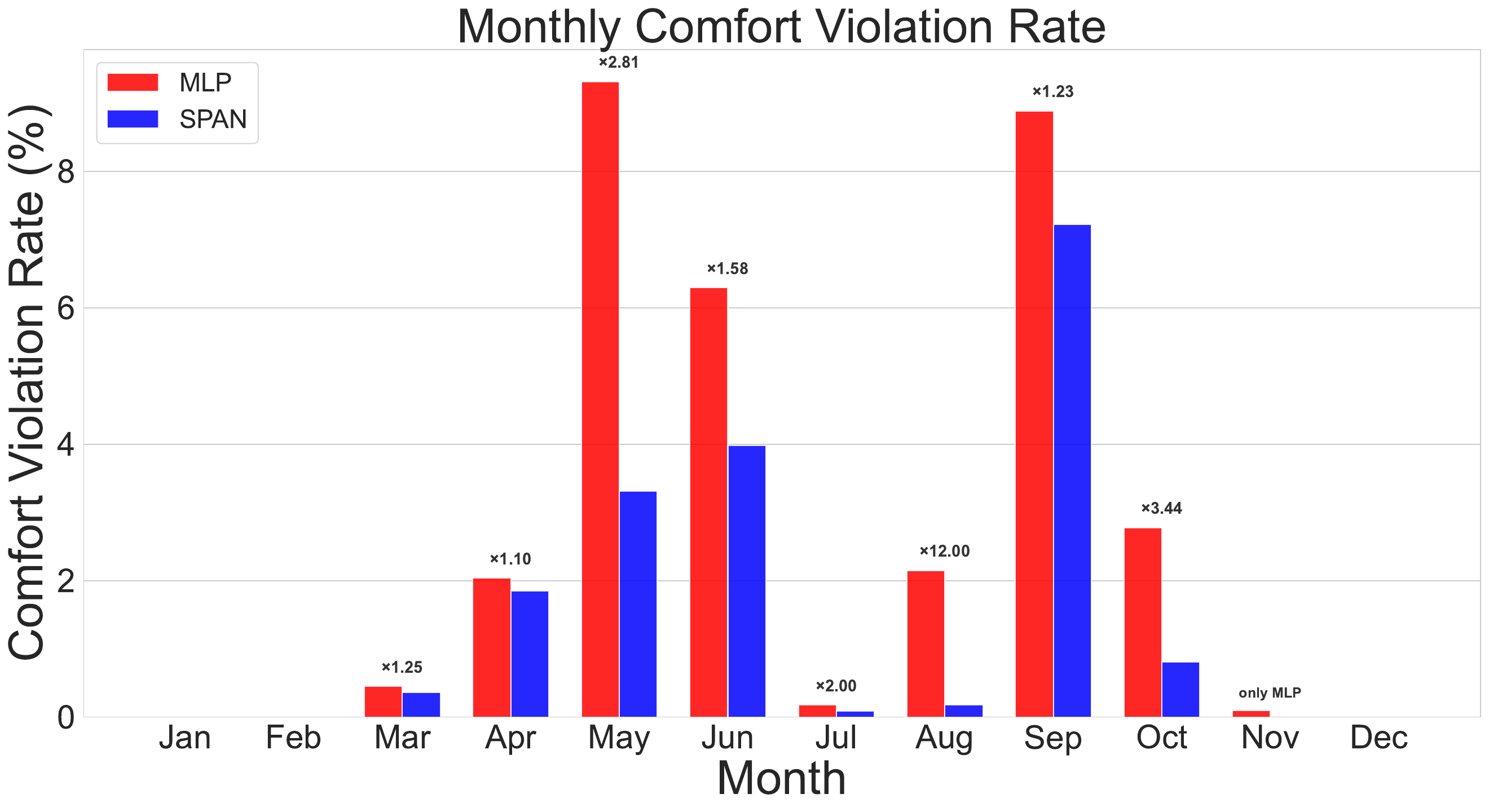}
        \caption{Monthly comfort violation rate (fraction of timesteps with indoor temperature $> 27^\circ$C or  $< 18^\circ$C) for SPAN and MLP, averaged over 5 seeds. Multipliers above MLP bars indicate how many times higher MLP's violation rate is relative to SPAN for that month.}
        \label{fig:sinergym_violations}
\end{figure}

This qualitative observation is confirmed quantitatively in Figure~\ref{fig:sinergym_violations}, which reports the monthly comfort violation rate the fraction of timesteps at which indoor temperature exceeds $27^\circ$C for both controllers. Neither controller ever violates the lower bound of $18^\circ$C. Violations are absent in January, February, and December, when low outdoor temperatures present no upper-bound thermal stress, and negligible in November (MLP only, 0.09\%). The violation pattern is strongly seasonal, peaking in the transition months of May, June, and September when rising or falling outdoor temperatures interact with building thermal mass to create the most variable indoor 
conditions. SPAN reduces comfort violations relative to MLP in every month where violations occur. The most striking reductions are in August ($\times$12.00: MLP 2.15\% vs.\ SPAN 0.18\%), May ($\times$2.81: MLP 9.32\% vs.\ SPAN 3.32\%), and October ($\times$3.44: MLP 2.78\% vs.\ SPAN 0.81\%). Even in September, where both controllers experience their highest absolute violation rates, SPAN reduces violations by a factor of $\times$1.23 (8.89\% vs.\ 7.22\%). 

%Across the full year, SPAN incurs comfort violations in 1.48\% of all timesteps compared to 2.69\% for MLP a 45\% reduction.

\begin{figure}[t]
    \centering
        \includegraphics[width=\linewidth]{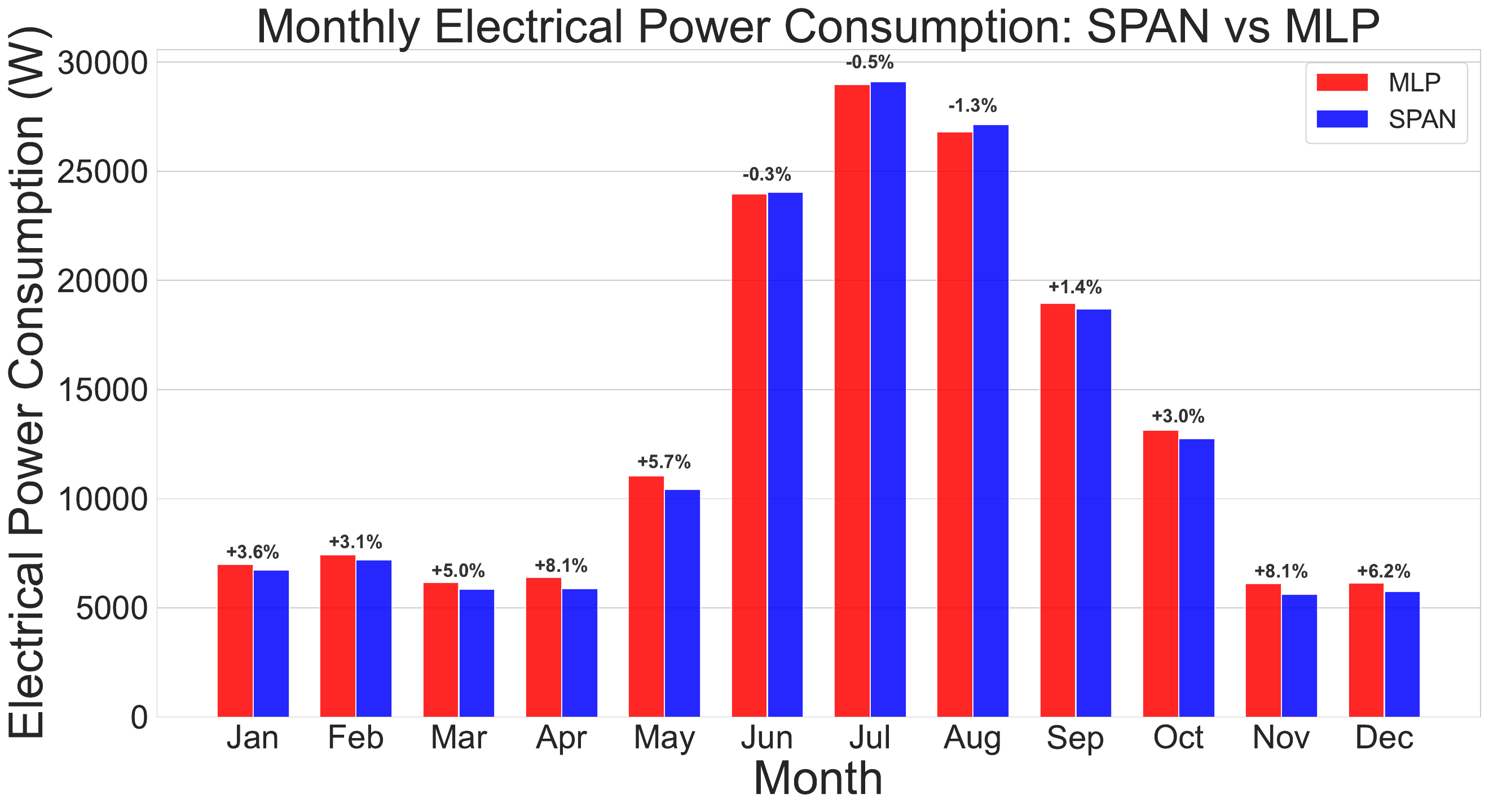}
        \caption{Monthly average HVAC electrical power consumption (W) for SPAN and MLP, averaged over 5 seeds. Percentages above each bar pair indicate the relative reduction achieved by SPAN with respect to MLP (positive values denote SPAN savings).}
        \label{fig:sinergym_power}
\end{figure}

Figure~\ref{fig:sinergym_power} reports monthly average HVAC electrical power consumption with the percentage reduction achieved by SPAN annotated above each bar pair. SPAN reduces energy consumption relative to MLP in nine of twelve months, with the largest savings in the shoulder seasons: April ($+8.1\%$), November ($+8.1\%$), May ($+5.7\%$), and December ($+6.2\%$). The sole exception is the peak summer period, where SPAN consumes marginally more energy than MLP in June ($-0.3\%$), July ($-0.5\%$), and August ($-1.3\%$). Crucially, the comfort violation data in Figure~\ref{fig:sinergym_violations} contextualises this peak-summer energy difference: SPAN's slightly higher power draw in July and August coincides with substantially fewer comfort violations in those months (June: $\times$1.58, July:$\times$2.00, August: $\times$12.00), indicating that SPAN is spending marginally more energy precisely to maintain thermal comfort under the most demanding cooling conditions rather than allowing violations.

\section{Discussion and Future Work} \label{sec:discussion}

This work demonstrates that SPAN is particularly effective in reinforcement learning regimes where samples, parameters, or training time are limited. Across classic control, continuous control, offline manipulation, and real-world datacenter HVAC optimization, SPAN consistently attains stronger policies with fewer environment interactions and remains effective under tight parameter budgets where standard MLPs often fail to learn. The Sinergym results extend this 
finding beyond synthetic benchmarks demonstrating that the architectural advantages observed in locomotion and control benchmarks transfer to physically grounded engineering problems with real thermodynamic dynamics and weather-driven non-stationarity. SPAN's local smoothness bias well suited to value functions and policies in physical systems, enables efficient representation with fewer parameters than globally connected MLPs. In addition, smooth derivatives reduce optimization variance, leading to more stable learning across random seeds. Unlike prior spline-based approaches relying on dense edge-wise constructions, SPAN’s low-rank tensor product structure achieves these benefits without incurring prohibitive computational overhead and accounting for convergence reliability through the expected training cost metric, SPAN is more computationally efficient than MLP. Despite these advantages, SPAN has limitations that bound its current scope. In sparse-reward settings, the local smoothness bias may fail as value function changes sharply near goal or end. In this current form SPAN is only designed as a feedforward state-action function approximator and does not naturally extend to architectures that rely on attention or recurrence as their primary representational component.Several research directions emerge naturally from the present work:  
\paragraph{Hybrid Architectures with Transformers.} Immediate architectural extension is a hybrid couples transformer backbone for temporal reasoning with SPAN head for the final value or policy output.  In sequence based offline RL settings such as Decision Transformer \citep{chen2021DT}, Trajectory Transformer \citep{janner2021TT}, transformer captures long horizon dependencies and SPAN exploits local smoothness for efficient approximation. 

\paragraph{Suspension Control and Design} Vehicle suspension control system present a control problem that must balance ride comfort, road holding and suspension travel in real time on limited hardware \citep{dridi2023}. Also multi-objective vehicle suspension system design \citep{desantanna2025morl} is another extension. Because the suspension dynamics are governed by physically smooth ODEs, SPAN should yield accurate policy surfaces. 

\paragraph{Industrial Process Control} PC-Gym \citep{bloor2024} provides a range of chemical and process engineering control problems, including continuously stirred tank reactors (CSTR), multistage extraction column and biofilm reactor. Applying SPAN to PC-Gym would extend present HVAC findings to a broader class of industrial process control tasks. 

\paragraph{Plasticity and Scaling.} A systematic study of SPAN's sacling behavior and its resistance to plasticity loss \citep{lyle2023plasticity} would clarify whether SPAN provides an architectural solution to plasticity degradation.

\section{Conclusion} \label{sec:conclusion}
Deploying reinforcement learning in resource-constrained environments, such as robotics control with limited samples, edge devices with strict memory budgets, production systems with uncertain training durations, and real-world engineering applications with measurable physical costs requires function approximators that learns efficiently. This work introduces SPAN, an architecture that addresses this need by aligning its inductive bias directly with the Lipschitz structure of RL value functions and policies.

It achieves a 30-50\% improvement in sample efficiency and 1.3-9$\times$ higher success rates in online benchmarks with 1.3-6.3$\times$ reduction in computational cost and on offline expert datasets, SPAN outperforms the MLP baseline by an average factor of 6.7$\times$. In the real-world datacenter application, SPAN reduces HVAC energy consumption in nine of twelve months while simultaneously achieving a 1.1-3.4 times reduction in thermal comfort violations, demonstrating that the architectural advantages observed in synthetic benchmarks transfer to physically grounded engineering control.

Taken together, these results establish SPAN as a principled and practically viable alternative to MLPs for resource-constrained RL. The architecture's local support, parameter efficiency, and convergence reliability make it particularly well-suited to settings where training budgets are limited, interactions are costly, or reliable anytime performance is required. More broadly, this work demonstrates that structured function approximators whose inductive bias is grounded in the mathematical properties of the functions being learned offer a productive and underexplored direction for improving the efficiency and reliability of deep reinforcement learning.

\nocite{langley00}
\clearpage

\bibliographystyle{cas-model2-names}
\bibliography{reference}
%%%%%%%%%%%%%%%%%%%%%%%%%%%%%%%%%%%%%%%%%%%%%%%%%%%%%%%%%%%%%%%%%%%%%%%%%%%%%%%
%%%%%%%%%%%%%%%%%%%%%%%%%%%%%%%%%%%%%%%%%%%%%%%%%%%%%%%%%%%%%%%%%%%%%%%%%%%%%%%
% APPENDIX
%%%%%%%%%%%%%%%%%%%%%%%%%%%%%%%%%%%%%%%%%%%%%%%%%%%%%%%%%%%%%%%%%%%%%%%%%%%%%%%
%%%%%%%%%%%%%%%%%%%%%%%%%%%%%%%%%%%%%%%%%%%%%%%%%%%%%%%%%%%%%%%%%%%%%%%%%%%%%%%
\clearpage
\appendix
\onecolumn
\section{Anytime Performance} ~\label{app:anytime}
This section presents comprehensive anytime performance results across all environments. The agent  performance is evaluated at five checkpoints during training: 10\% (100k), 25\% (250k), 50\% (500k), 75\% (750k), and 100\% (1M) of the total training budget steps. For each checkpoint, mean score across 20 seeds, standard deviation are reported.

Table~\ref{tab:anytime_classic} presents anytime performance for CartPole, LunarLander, and Acrobot. SPAN demonstrates consistent advantages across all checkpoints, with particularly pronounced benefits during early training phases when sample efficiency matters most.
\FloatBarrier
\begin{table}[H]
\centering
\small
\caption{Anytime performance on Classic Control environments. Mean score ± std across 20 seeds at each training checkpoint.}
\label{tab:anytime_classic}
\begin{tabular*}{\columnwidth}{@{\extracolsep{\fill}}llcccccc}
\toprule
Environment & Method & 10\% & 25\% & 50\% & 75\% & 95\% & 100\% \\
\midrule
\multirow{2}{*}{CartPole} 
& SPAN & 247±141 & \textbf{359±153} & \textbf{482±71} & \textbf{488±54} & \textbf{490±45} & \textbf{491±38} \\
\cmidrule{2-8}
& MLP & 247±134 & 221±125 & 429±123 & 449±128 & 461±113 & 464±112 \\
\midrule
\multirow{2}{*}{LunarLander} 
& SPAN & \textbf{-529±565} & \textbf{26±116} & \textbf{160±66} & \textbf{199±40} & \textbf{204±48} & \textbf{218±55} \\
\cmidrule{2-8}
& MLP & -785±711 & -14±65 & 108±52 & 178±39 & 196±33 & 195±39 \\
\midrule
\multirow{2}{*}{Acrobot} 
& SPAN & \textbf{-445±105} & \textbf{-223±178} & \textbf{-134±127} & \textbf{-112±94} & \textbf{-108±93} & \textbf{-104±93} \\
\cmidrule{2-8}
& MLP & -468±98 & -430±145 & -302±206 & -250±209 & -222±193 & -216±193 \\
\bottomrule
\end{tabular*}
\end{table}

On \textbf{CartPole}, SPAN shows stable learning from early checkpoints, while MLPs exhibit a notable performance dip at 25\% (221±125) before recovering. By 50\% of training, SPAN achieves 482±71 return compared to MLPs' 429±123, demonstrating both higher performance and substantially lower variance (71 vs 123 standard deviation). This stability advantage persists through final convergence, with SPAN maintaining tighter variance bounds throughout training.

\textbf{LunarLander} reveals SPAN's most dramatic early learning advantages. At 25\% of training budget, SPAN achieves positive returns (26±116) while MLPs remain slightly negative (-14±65). By 50\% checkpoint, SPAN reaches 160±66 already 73\% of its final performance compared to MLPs at 108±52 (55\% of final). This 40\% sample efficiency improvement at mid-training translates directly to reduced costs in sample expensive domains. Notably, SPAN continues improving through final convergence (218±55) while MLP performance plateaus around 75\% (178±39 $\rightarrow$ 195±39).

For \textbf{Acrobot}, SPAN demonstrates superior performance at every checkpoint with remarkably consistent improvement. The 25\% checkpoint comparison is particularly striking: SPAN achieves -223±178 versus MLPs at -430±145, a 63\% reduction in policy quality gap. Even at 10\%, SPAN shows modestly better mean performance with slightly higher variance, suggesting more exploratory but ultimately more successful learning dynamics. Final convergence maintains SPAN's advantage: -104±93 versus -216±193, representing both better final performance and 52\% variance reduction.

Table~\ref{tab:anytime_mujoco} presents anytime performance on Hopper, InvertedPendulum, walker2d and HalfCheetah. Across all MuJoCo tasks which is substantially more complex than classic control, SPAN demonstrates faster learning, more stable convergence, and dramatically reduced training variance. 
\begin{table}[!htbp]
\centering
\small
\caption{Anytime performance on MuJoCo environments. Mean score ± std across 20 seeds at each training checkpoint.}
\label{tab:anytime_mujoco}
\begin{tabular*}{\columnwidth}{@{\extracolsep{\fill}}llcccccc}
\toprule
Environment & Method & 10\% & 25\% & 50\% & 75\% & 95\% & 100\% \\
\midrule
\multirow{2}{*}{Hopper} 
& SPAN & \textbf{480±293} & \textbf{728±633} & \textbf{2213±1208} & \textbf{2371±1117} & \textbf{2366±1120} & \textbf{2647±971} \\
\cmidrule{2-8}
& MLP & 407±335 & 504±225 & 726±386 & 925±643 & 693±414 & 611±431 \\
\midrule
\multirow{2}{*}{InvertedPendulum} 
& SPAN & \textbf{728±386} & \textbf{953±212} & \textbf{926±227} & \textbf{1000±0} & \textbf{1000±0} & \textbf{975±113} \\
\cmidrule{2-8}
& MLP & 289±422 & 431±478 & 619±479 & 646±449 & 454±427 & 571±432 \\
\midrule
\multirow{2}{*}{HalfCheetah} 
& SPAN & \textbf{2808±380} & \textbf{4463±543} & \textbf{5308±458} & \textbf{5795±566} & \textbf{6099±542} & \textbf{6099±527} \\
\cmidrule{2-8}
& MLP & 1518±584 & 3229±437 & 4576±506 & 5118±596 & 5504±468 & 5544±659 \\
\midrule
\multirow{2}{*}{Walker2d} 
& SPAN & \textbf{316±147} & \textbf{557±159} & \textbf{890±737} & \textbf{1175±882} & \textbf{1312±878} & \textbf{1589±942} \\
\cmidrule{2-8}
& MLP & 256±224 & 509±334 & 513±268 & 531±181 & 563±255 & 582±222 \\
\bottomrule
\end{tabular*}
\end{table}

\textbf{Hopper} exemplifies SPAN's advantages on complex continuous control. From 10\% onward, SPAN maintains performance advantages that compound through training. The 50\% checkpoint reveals the critical difference: SPAN achieves 2213±1208 (88\% of final performance) while MLPs reach only 726±386 (29\% of final, actually exceeding final performance, indicating training instability). MLP performance degrades after 50\%, dropping from 925±643 at 75\% to 611±431 at convergence, suggesting policy collapse or catastrophic forgetting. In contrast, SPAN shows monotonic improvement with final performance of 2647±971 a 4.3× advantage over MLPs' 611±431. For practitioners, this means SPAN delivers usable policies at every checkpoint while MLPs risk complete failure. 

\textbf{InvertedPendulum} demonstrates SPAN's reliability through unprecedented stability. At 25\% budget, SPAN achieves 953±212 already 95\% of optimal (expert baseline: 1000) while MLPs reach only 431±478 (43\% of optimal). By 75\%, SPAN achieves perfect performance with zero variance (1000±0), maintaining this through 95\% before minor degradation at final checkpoint (975±113). MLPs never achieve consistent convergence, with final performance of 571±432 and persistent high variance indicating roughly half of training runs fail completely. This reliability difference is critical for production deployments where consistent performance across random seeds is essential.

\textbf{HalfCheetah} shows SPAN's sample efficiency advantages on high-dimensional continuous control. At just 10\% of training (100k steps), SPAN achieves 2808±380 versus MLPs' 1518±584 an 85\% performance advantage (46\% vs 27\% of respective final performance). This gap persists through mid-training: at 50\%, SPAN reaches 5308±458 (87\% of final) compared to MLPs' 4576±506 (83\% of final). While both methods eventually converge to respectable performance (SPAN: 6099±527, MLP: 5544±659), SPAN's 10\% advantage and consistently lower variance demonstrate its practical value for resource-constrained training scenarios. 

\textbf{Walker2d} reveals a subtler failure mode of MLP: MLP learning plateaus early, progressing only from 509±334 at 25\% to 582±222 at convergence a mere 14\% improvement over the final 75\% of training. SPAN, by contrast, improves continuously across all checkpoints, reaching 1589±942 at convergence, a 2.73× advantage. The gap opens at 50\% (890±737 vs.\ 513±268) and never closes, confirming that SPAN's local smoothness bias sustains meaningful policy improvement where MLP stagnates.

\subsection{Key Insights} These anytime performance results reveal three critical patterns. First, SPAN provides substantially better policies when training terminates early: at 25\% budget, SPAN shows 24-71\% better performance across environments, translating to usable controllers versus non functional policies. Second, SPAN exhibits dramatically reduced variance, particularly visible in InvertedPendulum (1000±0 vs 646±449 at 75\%) and Acrobot (-104±93 vs -216±193 final), indicating reliable convergence across random seeds. Third, MLP training on Hopper and Walker2d shows instability performance degradation after mid-training or high variance indicating frequent failures while SPAN maintains monotonic or stable improvement. For practitioners facing uncertain training budgets, hardware constraints, or requiring reliable convergence, these anytime performance characteristics make SPAN substantially more practical than standard MLPs despite comparable final performance in best case scenarios.

% Appendix B: Training Details and Hyperparameters

\section{Training Details and Hyperparameters}

\label{app:training_details}

This section provides complete hyperparameter configurations for reproducibility.
Table~\ref{tab:training_settings} summarizes total steps and number of seeds for training and ablation studies for each environment.

\begin{table}[h]
\centering
\small
\caption{Training and ablation settings. Steps and seeds refer to independent training runs.}
\label{tab:training_settings}
\begin{tabular*}{\columnwidth}{@{\extracolsep{\fill}}llcccc}
\toprule
 &  & \multicolumn{2}{c}{\textbf{Train}} & \multicolumn{2}{c}{\textbf{Ablation}} \\
\cmidrule(lr){3-4} \cmidrule(lr){5-6}
\textbf{Domain} & \textbf{Environment} & \textbf{Steps} & \textbf{Seeds} & \textbf{Steps} & \textbf{Seeds} \\
\midrule
Classic Control & CartPole-v1         & 500k & 20 & 300k & 5 \\
Classic Control & Acrobot-v1          & 500k & 20 & 300k & 5 \\
Box2D           & LunarLander-v3      & 1M   & 20 & 1M   & 5 \\
\midrule
                & InvertedPendulum-v5 & 1M   & 20 & 400k & 5 \\
MuJoCo          & Hopper-v5           & 1M   & 20 & 400k & 5 \\
                & HalfCheetah-v5      & 1M   & 20 & 400k & 5 \\
                & Walker2d-v5         & 1M   & 20 & 400k & 5 \\
\midrule
Application     & 2ZoneDataCenterHVAC & 525.6k &5 & - & -\\
\bottomrule
\end{tabular*}
\end{table}

Table~\ref{tab:common_hyperparams} shows hyperparameters across environments within each domain. Classic Control uses PPO while MuJoCo and HVAC Application uses SAC. And for offline learning IQL algorithm is used.

\begin{table}[h]
\centering
\small
\caption{Common training hyperparameters by learning paradigm}
\label{tab:common_hyperparams}
\begin{tabular*}{\columnwidth}{@{\extracolsep{\fill}}lccc}
\toprule
\textbf{Parameter} 
& \multicolumn{2}{c}{\textbf{Online RL}} 
& \textbf{Offline RL} \\
\cmidrule(lr){2-3}
& \textbf{Classic Control (PPO)} 
& \textbf{MuJoCo (SAC)} 
& \textbf{IQL} \\
\midrule
Total timesteps / iterations 
& $5 \times 10^5$ 
& $1 \times 10^6$ 
& 20,000 \\
Batch size 
& 1024 (rollout) 
& 128 
& 256 \\
Mini-batch size 
& 64 
& -- 
& -- \\
Update epochs 
& 4 
& 1 
& -- \\
Discount factor ($\gamma$) 
& 0.99 
& 0.99 
& 0.99 \\
GAE $\lambda$ 
& 0.95 
& -- 
& -- \\
Learning rate 
& $3 \times 10^{-4}$ 
& $3 \times 10^{-4}$ 
& $3 \times 10^{-4}$ \\
Value loss coefficient ($c_v$) 
& 0.5 
& -- 
& -- \\
Clip coefficient 
& 0.2 
& -- 
& -- \\
Max grad norm 
& 0.5 
& 10.0 
& -- \\
Replay buffer size 
& -- 
& $1 \times 10^6$ 
& Offline dataset \\
Random exploration steps 
& -- 
& 50,000 
& -- \\
Soft update ($\tau$) 
& -- 
& 0.005 
& 0.005 \\
Target entropy scale / temperature 
& -- 
& 1.0 
& 3.0 \\
Expectile ($\tau_{\text{exp}}$) 
& -- 
& -- 
& 0.7 \\
\midrule
Evaluation interval 
& 5,000 steps 
& 5,000 steps 
& -- \\
Evaluation episodes 
& 30 
& 30 
& 100 \\
Number of seeds 
& 20 
& 20 
& 5 \\
\bottomrule
\end{tabular*}
\end{table}

The number of parameters for SPAN and MLP architectures in online reinforcement learning is summarized in Table~\ref{tab:model_hyperparams}, with Table~\ref{tab:model_hyperparams_offline} presenting the values for offline reinforcement learning.

\begin{table*}[!htbp]
\centering
\small
\caption{Environment specific parameters}
\label{tab:model_hyperparams}
\begin{tabular*}{\textwidth}{@{\extracolsep{\fill}}lccccc|cc}
\toprule
\textbf{Environment} & \textbf{MLP Actor} & \textbf{MLP Critic} & \textbf{nmodes} & \textbf{nelems} & \textbf{degree} & \textbf{MLP Params} & \textbf{SPAN Params} \\
\midrule
\multicolumn{8}{l}{\textit{Classic Control}} \\
CartPole-v1         & (4, 3)   & (4, 4)   & 1  & 2 & 1 & 88   & 86   \\
Acrobot-v1          & (13, 13) & (13, 13) & 6  & 2 & 1 & 602  & 472  \\
LunarLander-v3      & (18, 17) & (20, 19) & 11 & 2 & 1 & 1156 & 1084 \\
\midrule
\multicolumn{8}{l}{\textit{MuJoCo}} \\
InvertedPendulum-v5 & (6, 5)   & (6, 7)   & 2  & 2 & 2 & 256  & 260  \\
Hopper-v5           & (25, 24) & (29, 29) & 10 & 8 & 2 & 3744 & 3799 \\
HalfCheetah-v5      & (21, 21) & (28, 28) & 10 & 4 & 2 & 4130 & 4147 \\
Walker2d-v5         & (23, 23) & (30, 30) & 10 & 6 & 2 & 4616 & 4623 \\
\midrule
\multicolumn{8}{l}{\textit{Application}} \\
2ZoneDataCenterHVAC & (52,51)  & (54,53)  & 10 & 8 & 2 & 15785 & 15859\\
\bottomrule
\end{tabular*}
\end{table*}

\begin{table*}[!htbp]
\centering
\small
\caption{Environment specific parameters}
\label{tab:model_hyperparams_offline}
\begin{tabular*}{\textwidth}{@{\extracolsep{\fill}}lcccccc|cc}
\toprule
\textbf{Environment} & \textbf{MLP Actor} & \textbf{MLP Critic} & \textbf{MLP Value} & \textbf{nmodes} & \textbf{nelems} & \textbf{degree} & \textbf{MLP Params} & \textbf{SPAN Params} \\
\midrule
\multicolumn{9}{l}{\textit{Adroit - Door}} \\
Expert & (36,36) & (64,64) & (45,45) & 15 & 4 & 2 & 17337 & 17449 \\
Cloned & (36,36) & (64,64) & (45,45) & 15 & 4 & 2 & 17337 & 17449 \\
Human & (36,36) & (64,64) & (45,45) & 15 & 4 & 2 & 17337 & 17449 \\
\midrule
\multicolumn{9}{l}{\textit{Adroit - Hammer}} \\
Expert & (41,41) & (68,68) & (50,50) & 15 & 4 & 2 & 20509 & 20448 \\
Cloned & (41,41) & (68,68) & (50,50) & 15 & 4 & 2 & 20509 & 20448 \\
Human & (41,41) & (68,68) & (50,50) & 15 & 4 & 2 & 20509 & 20448 \\
\midrule
\multicolumn{9}{l}{\textit{Adroit - Pen}} \\
Expert & (41,41) & (66,66) & (50,50) & 15 & 4 & 2 & 19634 & 19469 \\
Cloned & (41,41) & (66,66) & (50,50) & 15 & 4 & 2 & 19634 & 19469 \\
Human & (41,41) & (66,66) & (50,50) & 15 & 4 & 2 & 19634 & 19469 \\
\midrule
\multicolumn{9}{l}{\textit{Adroit - Relocate}} \\
Expert & (35,35) & (66,66) & (45,45) & 15 & 4 & 2 & 17845 & 17909 \\
Cloned & (35,35) & (66,66) & (45,45) & 15 & 4 & 2 & 17845 & 17909 \\
Human & (35,35) & (66,66) & (45,45) & 15 & 4 & 2 & 17845 & 17909 \\
\bottomrule
\end{tabular*}
\end{table*}

\section{Complete Ablation Studies}
\label{app:ablation}
Systematic ablation studies are conducted across CartPole, InvertedPendulum, Walker2d, to understand the impact of SPAN's architectural hyperparameters: \textit{nmodes} (number of tensor product modes), \textit{nelems} (number of B-spline elements per dimension), and \textit{degree} (polynomial degree of B-splines).

\textbf{CartPole-v1} demonstrates remarkable stability across all hyperparameter ranges. The environment maintains optimal return ($\sim$500) across nmodes $\in [1,9]$, nelems $\in [2,5]$, and degree $\in [1,4]$, with flat response curves indicating complete saturation at minimal capacity. This suggests that even the smallest SPAN configurations provide sufficient representational power for simple control tasks.

\textbf{InvertedPendulum-v5} reveals critical threshold effects in architectural capacity. Configurations with nmodes=2 and nelems=2 achieve near-optimal performance ($\sim$1000 return), while minimal settings (nmodes=1, nelems=1, degree=1) cause catastrophic failure with returns dropping to 400-600. Notably, once the minimum complexity threshold is exceeded, performance saturates immediately with no additional benefit from increased capacity, highlighting a sharp phase transition in model expressiveness.

\textbf{Walker2d-v5} exhibits sensitivity to both capacity parameters, with high variance reflecting the task's demand for sustained bipedal locomotion stability. Increasing \texttt{nmodes} from 8 to 16 yields a clear improvement in median return (from $\sim$1000 to $\sim$2500), with a further modest gain at 24, suggesting that representational rank is an important driver of policy quality for this environment. The \texttt{nelem} parameter shows a similarly strong effect: \texttt{nelem}=2 produces highly variable and often near-zero returns, while \texttt{nelem}=6 and \texttt{nelem}=8 achieve substantially higher and more consistent performance (median $\sim$3000 return), indicating that sufficient resolution in the spline basis is critical for capturing the complex dynamics of bipedal locomotion.

%\textbf{Summary.} Ablation studies reveal a clear pattern: simple tasks (CartPole, InvertedPendulum) are insensitive to hyperparameter choices once minimal capacity thresholds are met, while complex tasks (LunarLander, Hopper) benefit from higher capacity configurations (nmodes $\geq$ 10, nelems $\geq$ 6 for MuJoCo). Critically, once capacity thresholds are exceeded, performance stabilizes across wide hyperparameter ranges, demonstrating SPAN's practical robustness and eliminating the need for extensive environment-specific tuning.

\begin{figure*}[!htbp]
\centering
\begin{subfigure}[b]{0.96\textwidth}
    \includegraphics[width=\textwidth]{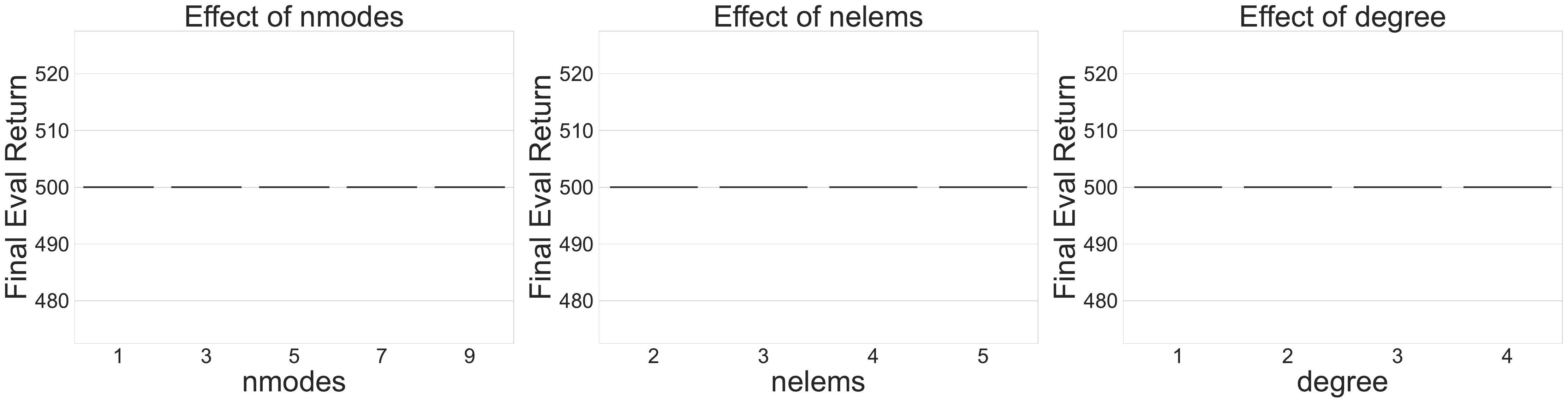}
    %\caption{CartPole}
    \label{fig:cartpole_ablation}
\end{subfigure}
\caption{Ablation study on CartPole-v1 showing performance vs. nmodes, nelems, and degree.}
\label{fig:cartpole_ablation}
\end{figure*}

\begin{figure*}[t]
\centering
\begin{subfigure}[b]{0.96\textwidth}
    \includegraphics[width=\textwidth]{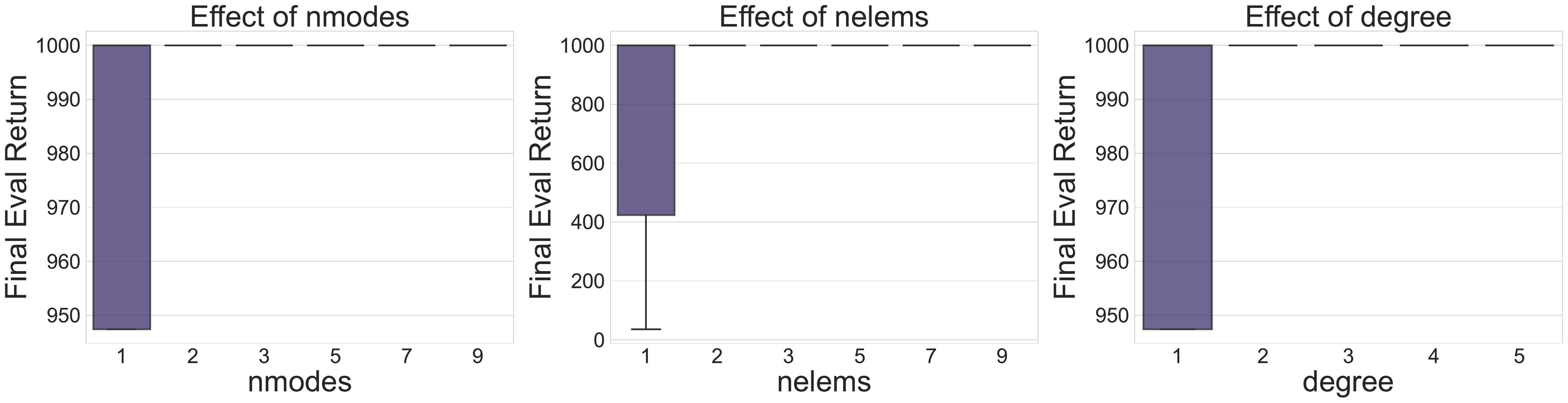}
    %\caption{CartPole}
    \label{fig:Invertedpendulum_ablation}
\end{subfigure}
\caption{Ablation study on InvertedPendulum-v5 showing performance vs. nmodes, nelems, and degree.}
\label{fig:Invertedpendulum_ablation}
\end{figure*}

\begin{figure*}[t]
\centering
\begin{subfigure}[b]{0.96\textwidth}
    \includegraphics[width=\textwidth]{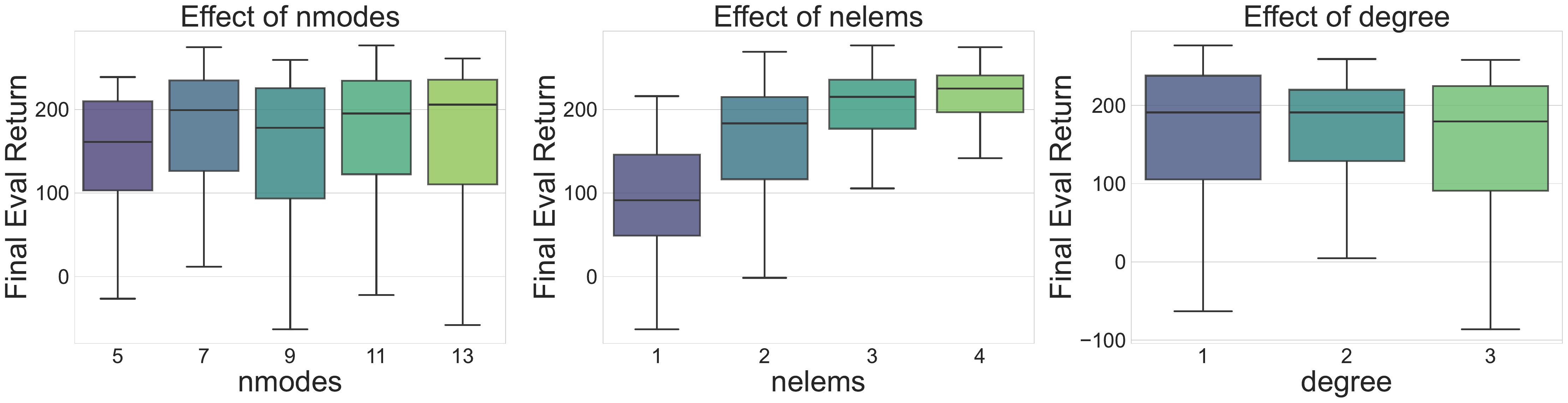}
    %\caption{CartPole}
    \label{fig:LunarLander_ablation}
\end{subfigure}
\caption{Ablation study on LunarLander-v3 showing performance vs. nmodes, nelems, and degree.}
\label{fig:LunarLander_ablation}
\end{figure*}

\begin{figure*}[!htbp]
\centering
\begin{subfigure}[b]{0.96\textwidth}
    \includegraphics[width=\textwidth]{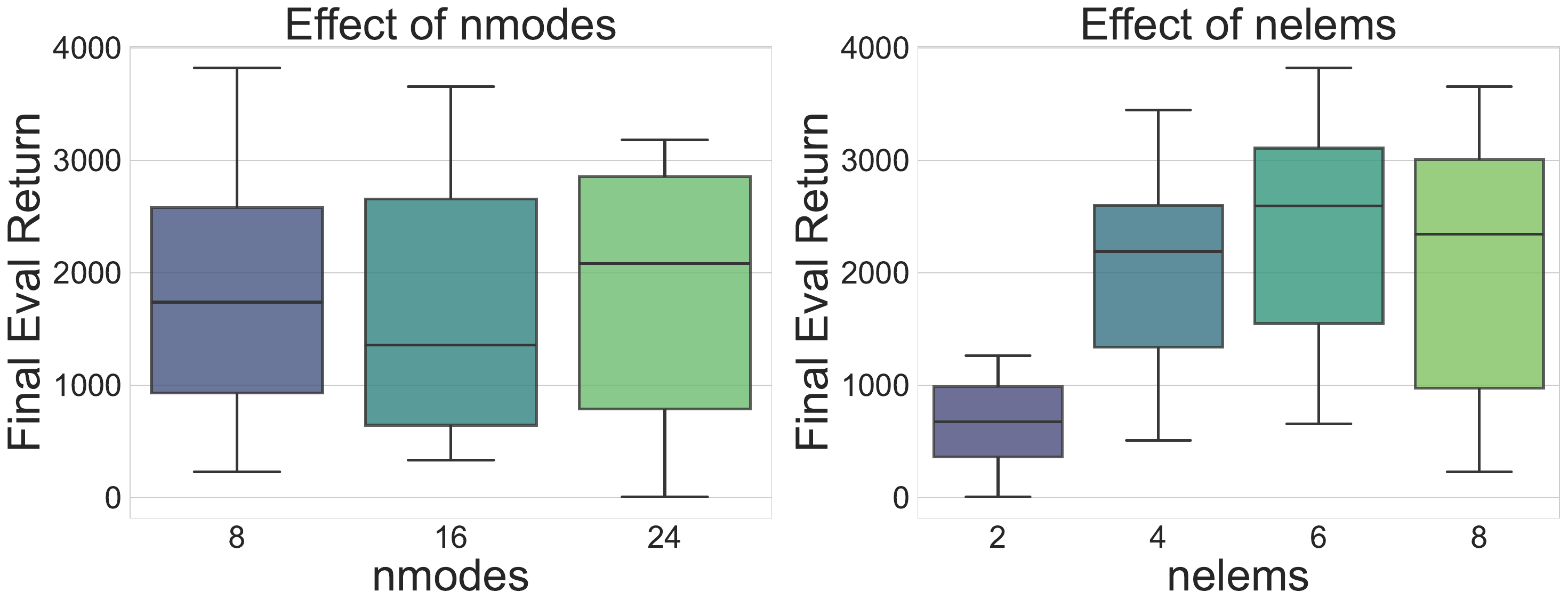}
    %\caption{CartPole}
    \label{fig:walker_ablation}
\end{subfigure}
\caption{Ablation study Walker-2D showing performance vs. nmode and nelem}
\label{fig:walker_ablation}
\end{figure*}

% ============================================================
% Appendix E — Statistical Evaluation Details
% ============================================================

\section{Full Wall-Clock and Expected Cost Breakdown}
\label{app:statistical}

This appendix provides the full wall clock and expected training cost breakdown across all performance thresholds. Table~\ref{tab:wallclock_full} extends Table~\ref{tab:compute} in Section~\ref{sec:compute} to report wall-clock time, success rate, and expected training cost at all five performance thresholds (25\%, 50\%, 75\%, 95\%, 100\%) for each environment. This provides a complete picture of how the computational advantage of SPAN evolves as a function of the target performance level. 

\begin{table}[h]
\centering
\small
\caption{Complete wall-clock time, success rate (SR), and expected training cost
($\mathcal{C}_{\mathrm{expected}} = T / \mathrm{SR}$) at all five performance
thresholds. `` - '' indicates zero seeds achieved the threshold within the training
budget (expected cost is infinite). All wall-clock times in minutes. SPS measurements
are constant per architecture and reported in Table~\ref{tab:compute}.}
\label{tab:wallclock_full}
\begin{tabular*}{\columnwidth}{@{\extracolsep{\fill}}llcccccccc}
\toprule
& &
\multicolumn{2}{c}{\textbf{SR}} &
\multicolumn{2}{c}{$T$ \textbf{(min)}} &
\multicolumn{2}{c}{$\mathcal{C}_{\mathrm{expected}}$ \textbf{(min)}} &
\multicolumn{2}{c}{\textbf{Ratio (MLP/SPAN)}} \\
\cmidrule(lr){3-4}\cmidrule(lr){5-6}\cmidrule(lr){7-8}\cmidrule(lr){9-10}
\textbf{Env.} & \textbf{Thresh.} &
SPAN & MLP & SPAN & MLP & SPAN & MLP & SR & $\mathcal{C}$ \\
\midrule
\multirow{5}{*}{CartPole}
 & 25\%  & 100\% & 95\%  & $2.5$    & $1.0$    & $2.5$    & $1.1$     & 1.05 & 0.44 \\
 & 50\%  & 100\% & 95\%  & $5.0$    & $4.1$    & $5.0$    & $4.3$     & 1.05 & 0.86 \\
 & 75\%  & 95\%  & 95\%  & $6.8$    & $9.7$    & $7.2$    & $10.2$    & 1.00 & 1.42 \\
 & 95\%  & 95\%  & 90\%  & $9.9$    & $10.5$   & $10.4$   & $11.7$    & 1.06 & 1.12 \\
 & 100\% & 95\%  & 75\%  & $13.7$   & $14.5$   & $14.4$   & $19.3$    & 1.27 & 1.34 \\
\midrule
\multirow{5}{*}{Acrobot}
 & 25\%  & 100\% & 100\% & $8.0$    & $10.8$   & $8.0$    & $10.8$    & 1.00 & 1.35 \\
 & 50\%  & 100\% & 70\%  & $8.9$    & $10.8$   & $8.9$    & $15.4$    & 1.43 & 1.73 \\
 & 75\%  & 95\%  & 60\%  & $9.4$    & $12.2$   & $9.9$    & $20.3$    & 1.58 & 2.05 \\
 & 95\%  & 90\%  & 60\%  & $11.0$   & $13.5$   & $12.2$   & $22.5$    & 1.50 & 1.84 \\
 & 100\% & 90\%  & 60\%  & $12.3$   & $14.0$   & $13.7$   & $23.3$    & 1.50 & 1.70 \\
\midrule
\multirow{5}{*}{LunarLander}
 & 25\%  & 100\% & 100\% & $26.7$   & $24.2$   & $26.7$   & $24.2$    & 1.00 & 0.91 \\
 & 50\%  & 100\% & 100\% & $27.7$   & $33.1$   & $27.7$   & $33.1$    & 1.00 & 1.19 \\
 & 75\%  & 95\%  & 95\%  & $42.1$   & $43.2$   & $44.3$   & $45.5$    & 1.00 & 1.03 \\
 & 95\%  & 95\%  & 75\%  & $54.8$   & $52.2$   & $57.7$   & $69.6$    & 1.27 & 1.21 \\
 & 100\% & 90\%  & 60\%  & $60.5$   & $56.3$   & $67.2$   & $93.8$    & 1.50 & 1.40 \\
\midrule
\multirow{5}{*}{\shortstack[l]{Inverted\\Pendulum}}
 & 25\%  & 100\% & 100\% & $35.0$   & $53.9$   & $35.0$   & $53.9$    & 1.00 & 1.54 \\
 & 50\%  & 100\% & 100\% & $36.8$   & $53.9$   & $36.8$   & $53.9$    & 1.00 & 1.46 \\
 & 75\%  & 100\% & 100\% & $36.8$   & $53.9$   & $36.8$   & $53.9$    & 1.00 & 1.46 \\
 & 95\%  & 100\% & 100\% & $36.8$   & $53.9$   & $36.8$   & $53.9$    & 1.00 & 1.46 \\
 & 100\% & 100\% & 100\% & $36.8$   & $53.9$   & $36.8$   & $53.9$    & 1.00 & 1.46 \\
\midrule
\multirow{5}{*}{HalfCheetah}
 & 25\%  & 100\% & 100\% & $83.1$   & $69.7$   & $83.1$   & $69.7$    & 1.00 & 0.84 \\
 & 50\%  & 100\% & 100\% & $133.0$  & $134.6$  & $133.0$  & $134.6$   & 1.00 & 1.01 \\
 & 75\%  & 100\% & 100\% & $271.5$  & $338.2$  & $271.5$  & $338.2$   & 1.00 & 1.25 \\
 & 95\%  & 90\%  & 35\%  & $653.8$  & $592.7$  & $726.4$  & $1693.4$  & 2.57 & 2.33 \\
 & 100\% & 60\%  & 10\%  & $773.5$  & $627.6$  & $1289.2$ & $6276.0$  & 6.00 & 4.87 \\
\midrule
\multirow{5}{*}{Hopper}
 & 25\%  & 95\%  & 90\%  & 198.7      & 129.5     & 209.2      & 143.9       & 1.06 & 0.69  \\
 & 50\%  & 90\%  & 10\%  & 304.5      & 212.8      & 338.3      & 2128       & 9.00 & 6.29  \\
 & 75\%  & 90\%  & 0\%   & 326.9      & -      & 363.3      & $\infty$  & -  & $\infty$  \\
 & 95\%  & 80\%  & 0\%   & 315.0      & -      & 393.8      & $\infty$  & -  & $\infty$  \\
 & 100\% & 80\%  & 0\%   & $315.0$  & -      & $393.8$  & $\infty$  & -  & $\infty$ \\
\midrule
\multirow{5}{*}{Walker2d}
 & 25\%  & 65\%  & 20\%  & $631.5$  & $388.2$  & $971.5$  & $1941.0$  & 3.25 & 2.00 \\
 & 50\%  & 55\%  & 0\%   & $737.7$  & -      & $1341.3$ & $\infty$  & -  & $\infty$ \\
 & 75\%  & 30\%  & 0\%   & $770.8$  & -      & $2569.3$ & $\infty$  & -  & $\infty$ \\
 & 95\%  & 0\%   & 0\%   & -      & -      & $\infty$ & $\infty$  & -  & -  \\
 & 100\% & 0\%   & 0\%   & -      & -      & $\infty$ & $\infty$  & -  & -  \\
\bottomrule
\end{tabular*}
\end{table}
\section{Hardware Specification} \label{app:hardware}

All experiments were conducted on a laptop computer with the following specifications:
\begin{itemize}
    \item \textbf{Processor:} Intel(R) Core(TM) i9-14900HX @ 2.20 GHz
    \item \textbf{RAM:} 32.0 GB (31.7 GB usable) @ 5600 MT/s
    \item \textbf{Graphics:} NVIDIA GeForce RTX 5060 (8 GB VRAM)
    %\item \textbf{Storage:} 954 GB SSD
    \item \textbf{Operating System:} Windows 11, 64-bit, x64-based processor
\end{itemize}

%% Bibliography 
%\bibliography{references}

\end{document}